\documentclass[
]{ceurart}

\usepackage{listings}
\usepackage{booktabs}
\usepackage{tabularx}
\usepackage{threeparttable}
\usepackage{subcaption}
\usepackage{wrapfig}
\usepackage{caption}
\usepackage{xparse}

\makeatletter
\@ifundefined{ccsdesc}{\newcommand{\ccsdesc}[2][]{}}{}
\@ifundefined{CCSXML}{%
  \newenvironment{CCSXML}{}{}%
}{}
\makeatother

\newif\ifprintccs
\printccsfalse

\newcommand{\CCSmetadata}[1]{\ifprintccs #1\fi}

\begin{document}

\copyrightyear{2026}
\copyrightclause{Copyright for this paper by its authors.
  Use permitted under Creative Commons License Attribution 4.0
  International (CC BY 4.0).}

\conference{Joint Proceedings of the ACM Intelligent User Interfaces (IUI) Workshops 2026, March 23-26, 2026, Paphos, Cyprus}

\title{Occlusion-Robust Multimodal Emotion Recognition in VR via Fusion of Facial Images and EMG}

\author[1]{Birgit Nierula}[%
orcid=0000-0003-2174-5872,
email=Birgit.Nierula@hhi.fraunhofer.de,
]
\cormark[1]
\fnmark[1]
\address[1]{Fraunhofer HHI, Berlin, Germany}

\author[1]{Karam Tomotaki-Dawoud}[%
orcid=0009-0006-2745-8312,
]
\fnmark[1]
\author[1,2]{Mert Akguel}
\address[2]{Humboldt University of Berlin, Berlin, Germany}
\author[1]{Mustafa Tevfik Lafci}
\author[1]{David Przewozny}
\author[1]{Anna Hilsmann}
\author[1,2]{Peter Eisert}
\author[1]{Sebastian Bosse}

\cortext[1]{Corresponding author.}
\fntext[1]{These authors contributed equally.}

\begin{abstract}
Head-mounted displays (HMDs) fundamentally limit emotion recognition in virtual reality (VR): by occluding the upper face, they render conventional image-based facial expression analysis incomplete, particularly for applications requiring real-time affective assessment.
We address this challenge by fusing lower-face video with facial electromyography (EMG) from the occluded upper face to classify seven emotional categories (six basic emotions plus neutral). We introduce a synchronized multimodal dataset from 20 participants, pairing lower-face video with seven-channel upper-face EMG elicited by validated emotion stimuli. Under subject-independent test, our proposed late-fusion architecture merging convolutional visual embeddings with RBF-kernel EMG representations achieves $51\%$ macro-F1, outperforming both image-only ($41\%$) and EMG-only ($43\%$) baselines. These results demonstrate that upper-face EMG provides robust complementary information under HMD-induced visual occlusion and establish a foundation for multimodal emotion recognition in naturalistic VR environments. This approach facilitates affect-adaptive applications, including communication training and therapeutic interventions. The dataset will be shared upon request under an ethical-use agreement.
\end{abstract}

\begin{keywords}
Virtual Reality \sep
Affective Computing \sep
Facial EMG \sep
Multimodal Fusion \sep
Emotion Recognition \sep
HMD Occlusion \sep
\end{keywords}


\maketitle

\CCSmetadata{%

\ccsdesc[500]{Computing methodologies~Machine learning}
\ccsdesc[500]{Human-centered computing~Virtual reality}
\ccsdesc[300]{Computing methodologies~Mixed / augmented reality}
}

\section{INTRODUCTION}
Virtual Reality (VR) systems have evolved beyond simple visual immersion to become platforms for human interaction and training. As these systems increasingly simulate complex social scenarios, they require mechanisms to understand and respond to users' emotional states in real-time. The immersive nature of VR enables the subjective experience of presence, the sensation of "being there"~\cite{SlaterWilbur1997FIVE}, which elicits authentic emotional responses comparable to real-world experiences~\cite{Felnhofer2015VRemotions, tian2021emotional}. This ability to induce genuine emotional responses while maintaining full control over the scenario~\cite{sanchez2005presence} makes VR a valuable tool for professional training applications, particularly in domains that require good communication skills, such as in healthcare~\cite{fernandez-alcantara_2025} or law enforcement~\cite{murtinger_2024}. In such scenarios, automatic recognition of trainees’ emotional expressions can provide objective feedback and enable adaptive role-play, supporting skill acquisition in emotionally demanding contexts.

Interpersonal communication has verbal and non-verbal aspects and one of the most important non-verbal ones is facial expressions, because they serve as one of the primary channels to convey emotional states~\cite{sagliano_2022} and are therefore relevant cues in a communication process. People are often unaware of their facial expressions, making assessment tools valuable for providing feedback and increasing self-awareness during communication.

Current approaches to emotion assessment in VR predominantly focus on psychophysiological measurements that capture cognitive or autonomic responses, such as changes in alpha frequency of the electroencephalogram~\cite{hofmann2021decoding}, heart rate variability~\cite{McCall_2015_Physiophenomenology, marin2021heart, nierula_2025}, or electrodermal activity~\cite{MarinMorales2020Sensors, llobera_2010, nierula_2025}. These physiological approaches have shown success across diverse VR applications, from phobia treatment protocols~\cite{Peperkorn_2014} to therapeutic interventions~\cite{maples2017treatment} and presence evaluation~\cite{diemer2015impact}. While these physiological signals provide valuable insights into arousal and valence of the user's experienced emotional state, they do not provide information about its expression as non-verbal communication cues. 

Implementing facial emotion recognition (FER) in VR poses the challenge that the head-mounted displays (HMDs) systematically occlude the upper facial region, including eyes, eyebrows and forehead that provide essential discriminative features for emotion classification. This occlusion fundamentally constrains the applicability of established camera-based facial emotion recognition methods~\cite{ortmann2023facial}, which traditionally depend on complete facial analysis to differentiate between subtle emotional distinctions, such as distinguishing surprise from fear, or anger from disgust.

The Facial Action Coding System (FACS) is a comprehensive, anatomically based framework developed by Paul Ekman and Wallace Friesen for systematically coding all observable facial movements~\cite{ekman1978manual}. It provides an objective, anatomy-grounded taxonomy for categorizing visible facial movements into discrete Action Units (AUs). While FACS traditionally relies on visual observation, facial electromyography (EMG) offers a complementary approach by directly measuring the underlying muscle activity of the AUs. 

In this study, we propose a novel multimodal approach that combines computer vision analysis of the visible lower face with facial EMG signals captured from the occluded upper regions. We systematically evaluate this fusion strategy for classifying seven distinct emotional states, the six basic emotions introduced by Ekman~\cite{Ekman_1992} plus a neutral baseline. Our contributions are threefold: (i) the first synchronized dataset combining external lower-face video with upper-face EMG under VR HMD occlusion; (ii) a systematic evaluation of multimodal late-fusion strategies including kernelized vs.\ direct EMG encodings; and (iii) demonstration of how multimodal fusion substantially outperforms unimodal baselines in this challenging setting.

\section{Related Work}
\label{Relatedwork}

Immersive virtual-reality (VR) experiences elicit genuine affect while preserving experimental control, thereby enabling ecologically valid emotion studies~\cite{MarinMorales2020Sensors,diemer2015impact}. Systematic reviews show a sharp rise in emotion-recognition studies that couple VR with physiological measures such as Electroencephalography (EEG), heart-rate variability or electro-dermal activity~\cite{MarinMorales2020Sensors}.  

The challenge of FER in VR contexts has inspired several innovative technical solutions to address the fundamental problem of upper-face occlusion: (i) Optical work-arounds repurpose inward-facing cameras, for example, peri-ocular IR images from eye-tracking cameras alone are sufficient to classify a subset of five emotive expressions~\cite{Hickson2019WACV}. (ii) A complementary optical approach places photo-reflective IR sensors in the visor to capture subtle skin deformations; \cite{Murakami2019SIGGRAPH} recognized basic expressions and demonstrated real-time avatar animation. (iii) A third line reconstructs the occluded region by enabling gaze-aware, photo-realistic facial re-enactment for HMD users~\cite{Thies2018FaceVR}, and generative RGB-D completion removes the headset and restores identity-preserving face geometry~\cite{Numan2021HMDRemoval}. Recent work further drives avatars from egocentric, headset-internal IR views in real time~\cite{Zhang2024REFA}.

Despite growing interest in VR-based emotion recognition, the availability of comprehensive multimodal datasets remains limited. Several notable contributions have begun addressing this gap. 
AVDOS-VR is a database of subjective and physiological measures to affective videos recorded during a remote VR-experiment. It contains continuous valence and arousal ratings, heart response and facial EMG~\cite{Gnacek2024SciData}. The emotiHeroVR Database contributes 3{,}556 labeled lower-face images together with 63 headset-internal expression activations from a Meta Quest Pro~\cite{Ortmann2024EmojiHeroVR} and was collected with the VR-game EmojiHeroVR. The game gives direct feedback in form of a rewarding tone whether a facial expression indicated by an emoji was detected or not. Prior to the game, participants were trained on how to optimally perform each emotional facial expression. EmojiHeroVR thus does not extract natural emotional facial expressions but instead facial expressions that the user has learned to be detected best by the system. VREED is a multimodal affective dataset of emotional states triggered via immersive 360° video-based virtual environments in an HMD. It consists of eye tracking, electrocardiography, and galvanic skin response, and subjective experience reports~\cite{luma2022vreed}. The HEADSET collection takes a different approach, capturing emotional expressions in a volumetric camera studio with selective HMD occlusion to systematically study emotion recognition under partial visibility constraints~\cite{Lohesara2023HEADSET}.

Facial surface EMG offers an elegant solution to occlusion by directly sensing muscle activity. The emteqPRO platform implements this approach by integrating 7 facial EMG channels with photoplethysmography and inertial measurement units within a VR-compatible form factor~\cite{Gjoreski2021Ubicomp, EmteqLabsPaper}. Using this mask,~\cite{kiprijanovska2022facial} achieved F1-macro scores up to 0.86 for five posed expressions under controlled conditions. However, their approach classified isolated facial movements rather than emotions: participants produced single-muscle actions (smile, frown, eyebrow raise, eye squeeze, and neutral), each corresponding to one primary muscle (zygomaticus major, corrugator supercilii, frontalis, orbicularis oculi). In contrast, emotions typically involve more than one muscle, even in this limited sensor setup (see for example that anger and happiness activate 2 sensors in table~\ref{tab:au_mapping}), making emotion recognition substantially more complex than detecting isolated facial movements. 
The same sensing technology has been successfully applied to continuous valence-arousal monitoring during affective VR video viewing~\cite{Gjoreski2022SciRep}.

The fusion of complementary sensor modalities helps overcoming individual sensor limitations. Early fusion schemes mixing forehead biopotentials with peripheral physiology improved robustness~\cite{Khezri2015Fusion}. In VR, multimodal pipelines that integrate cardiovascular, electro-dermal and inertial features outperform unimodal baselines and enable explainability~\cite{Polo2025CBM}. Deep fusion for VR emotion recognition has likewise shown benefits over single streams~\cite{indrasiri2024multimodal}. On the vision side, the EmojiHeroVR system reported gains when merging lower-face images with headset-internal expression activations~\cite{Ortmann2024EmojiHeroVR}.

In terms of application benefits, real-time inference can enable affect-adaptive interaction. A classic example is adapting the difficulty of \emph{Tetris} using EEG and peripheral signals to regulate boredom, engagement, and anxiety~\cite{Chanel2011SMC}.

\paragraph{Summary and novelty relative to prior EMG-in-VR work.}
Prior facial-EMG systems for VR demonstrate that upper-face EMG remains informative under HMD occlusion, but existing benchmarks often target (i) isolated facial muscle acitivity rather than multi-muscle emotion patterns, and/or (ii) continuous valence--arousal monitoring rather than discrete multi-class emotion recognition. In parallel, VR FER datasets and systems do not provide synchronized external video paired with upper-face EMG. In contrast, our work contributes (1) a synchronized dataset pairing external multi-view lower-face video with upper-face EMG under realistic HMD occlusion, (2) a systematic late-fusion evaluation for seven discrete emotion classes, and (3) an explicit comparison of direct EMG features versus kernelized EMG representations for multimodal fusion in a cross-subject setting.

\begin{figure}[t] 
  \centering
  \includegraphics[width=0.6\linewidth]{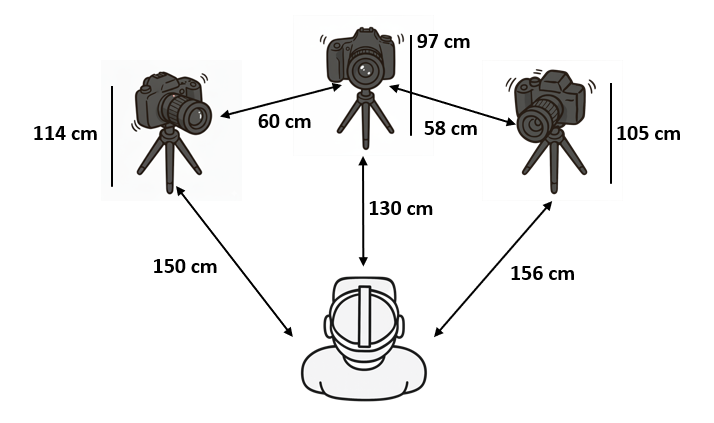}
  \caption{Experimental setup with three synchronized industrial monochrome cameras 
    (Sony IMX547 CMOS sensor, $2448{\times}2048$\,px, 22\,fps) positioned at $0^\circ$ and $\pm 30^\circ$ 
    relative to the sagittal plane. The monochrome design was selected for its higher sensitivity and 
    stable performance under the low-light conditions of the immersive VR environment.}
  \label{fig:camera_setup}
\end{figure}

\begin{figure}[t]
  \centering
    \centering
    \includegraphics[width=0.25\linewidth]{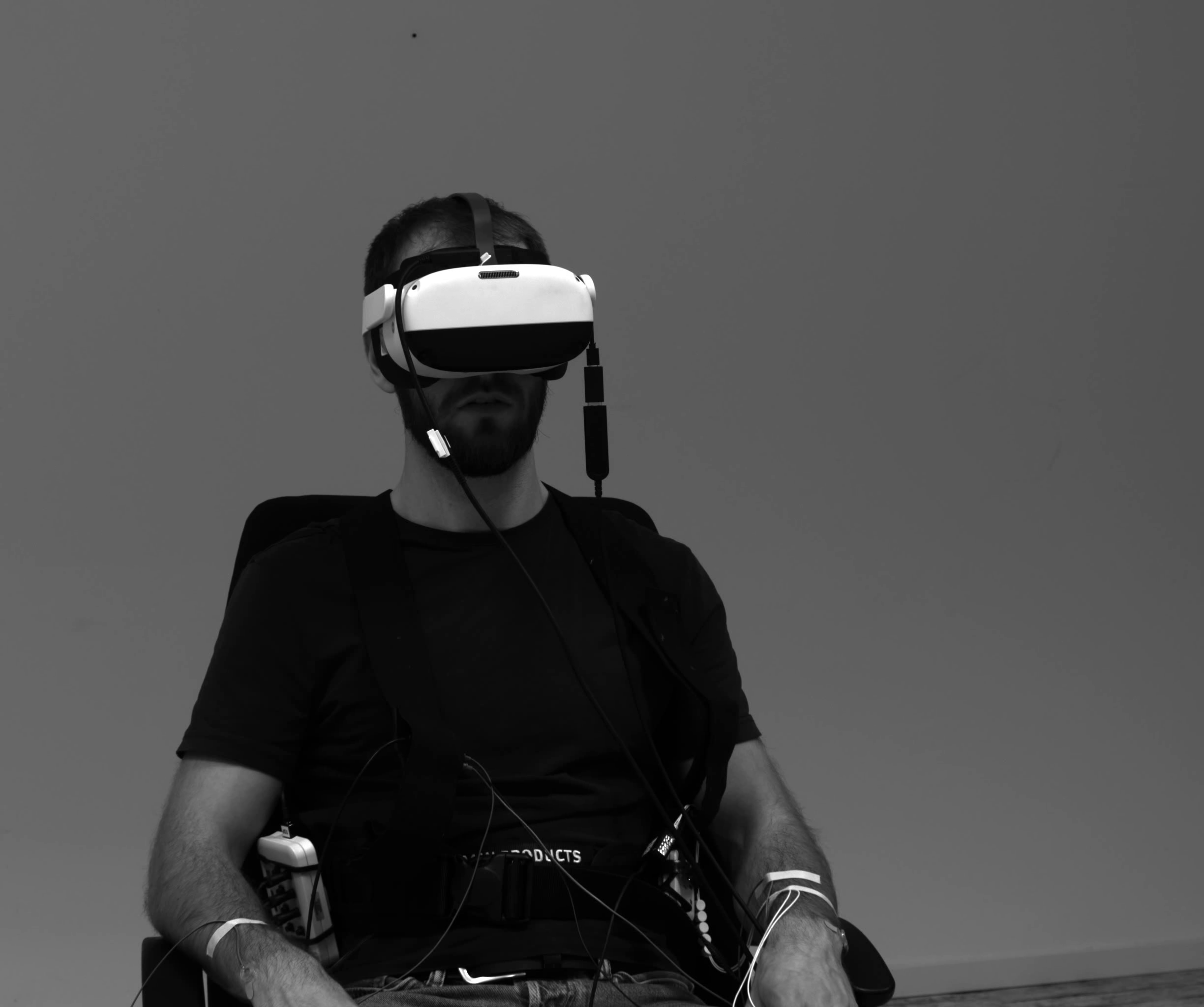}
    \includegraphics[width=0.25\linewidth]{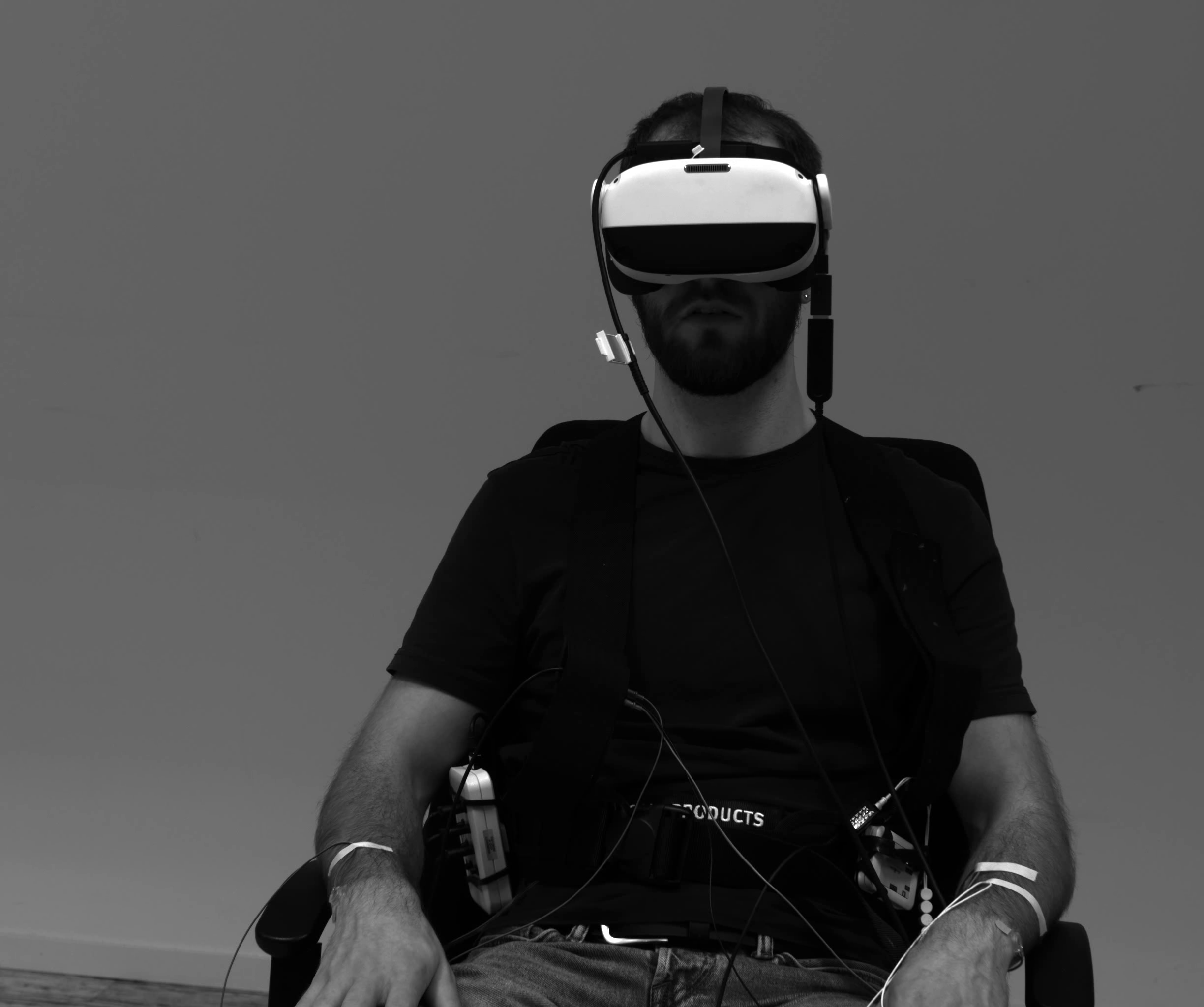}
    \includegraphics[width=0.25\linewidth]{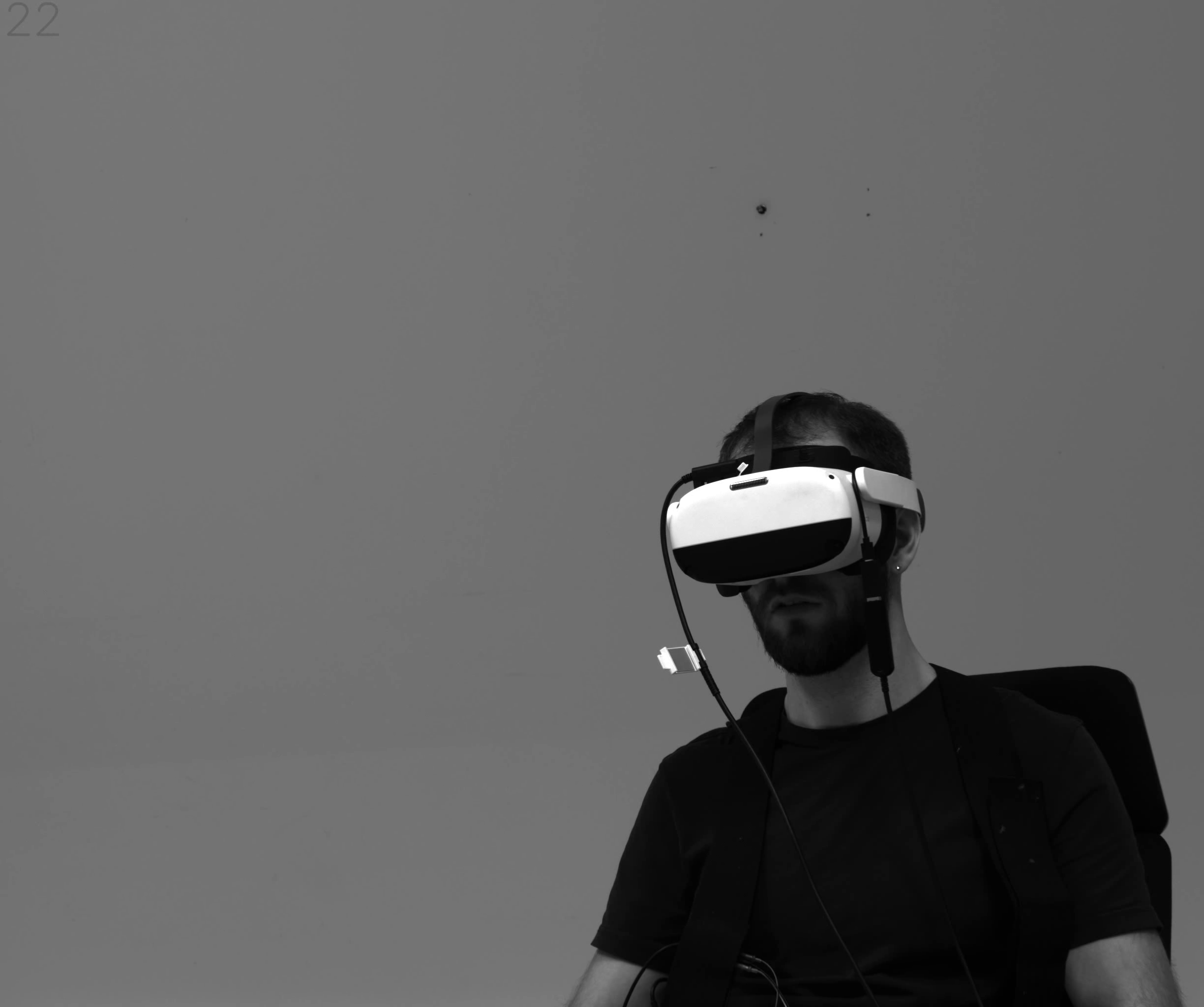}

  \vspace{0.5em} 

  
    \centering
    \includegraphics[width=0.25\linewidth]{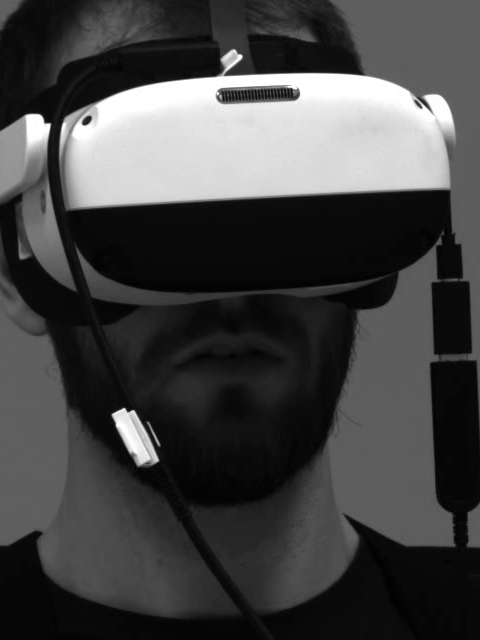}
    \includegraphics[width=0.25\linewidth]{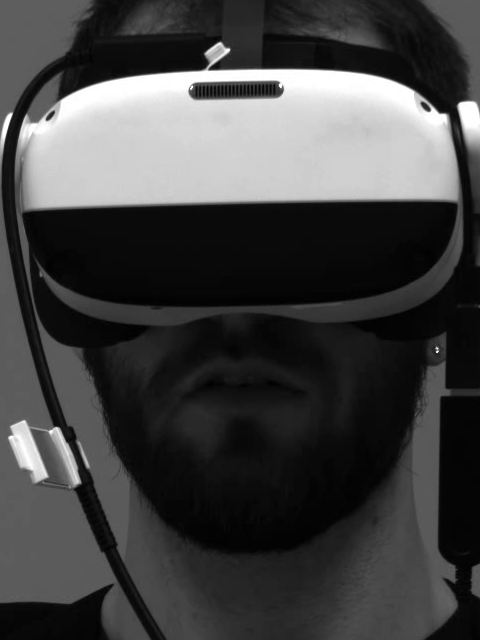}
    \includegraphics[width=0.25\linewidth]{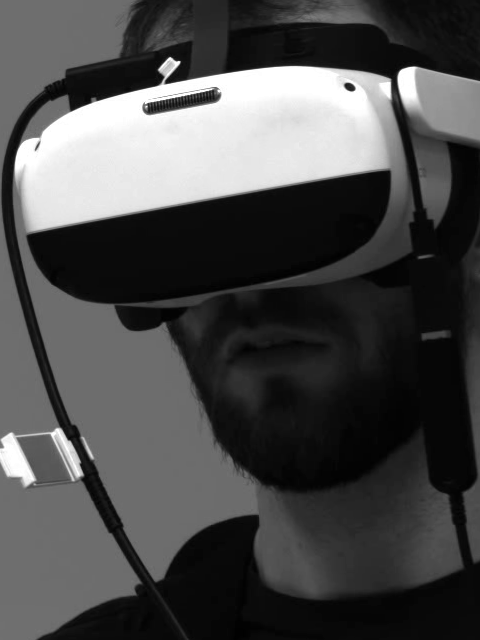}

  \caption{Example frames from each monochrome camera view (left, center, right): raw frames (top) and YOLO Pose lower-face crops (bottom) at the same timestamp.}
  \label{fig:raw_yolo_examples}
\end{figure}

\section{Dataset}
\label{Dataset}

We introduce and will contribute a new synchronized multimodal VR dataset pairing lower-face video with seven-channel upper-face EMG under HMD occlusion (details on access in Sec.~\ref{datasetAvailability}).

\paragraph{Participants.} 
Twenty volunteers (10 female, mean age = 27.4 years, SD = 6.7 years) were recruited. All participants reported normal or corrected-to-normal vision. 

\paragraph{Apparatus.} 
Stimuli were presented via a Pico Neo 3 HMD (per-eye resolution: $2296{\times}2408$ pixels, 72\,Hz refresh rate, $\sim98^\circ$ FoV). The HMD was equipped with an EmteqPRO open-face mask~\cite{EmteqLabsPaper}, which synchronously recorded seven-channel facial EMG from the following muscle groups: bilateral frontalis, orbicularis oculi, zygomaticus major, and central corrugator supercilii. These muscles correspond to key FACS AUs (Table~\ref{tab:au_mapping}), providing physiological signals that complement vision-based AU features.

\begin{table}[t]
\centering
\caption{Facial EMG channels, corresponding FACS AUs, and typical emotion associations.}
\begin{threeparttable}
\begin{tabularx}{\linewidth}{l X X}
\toprule
\textbf{Muscle (channel)} & \textbf{Primary AU(s)} & \textbf{Often seen in} \\
\midrule
Corrugator supercilii & AU4 (Brow Lowerer) & Anger, Sadness, Fear \\
Frontalis$^*$ (medialis)  & AU1 (Inner Brow Raiser) & Sadness, Fear, Surprise \\
Frontalis$^*$ (lateralis) & AU2 (Outer Brow Raiser) & Surprise \\
Orbicularis oculi     & AU6 (Cheek Raiser), AU7 (Lid Tightener) & Happiness, Anger \\
Zygomaticus major     & AU12 (Lip Corner Puller) & Happiness \\
\bottomrule
\end{tabularx}
\begin{tablenotes}
\footnotesize
\item[*] Note the EMG sensor cannot distinguish medial and lateral frontalis.
\end{tablenotes}
\end{threeparttable}
\label{tab:au_mapping}
\end{table}

Lower-face video was recorded with three synchronized industrial monochrome cameras (resolution $2448{\times}2048$\,px, 22\,fps) based on the Sony IMX547 CMOS sensor, placed at $0^\circ$ and $\pm 30^\circ$ relative to the sagittal plane (Fig.~\ref{fig:camera_setup}). 
The monochrome design was chosen for its higher sensitivity and stable performance under the low-light conditions of our immersive VR setting. This setup captured lower-face images from three complementary viewpoints 
(Fig.~\ref{fig:raw_yolo_examples}, top row).

\paragraph{Stimuli.} 
We used 35 facial expression images from the NimStim database~\cite{NimStim_dataset}, a validated stimulus set for emotion research. Five exemplars were selected from each of seven categories (happy, sad, angry, fearful, surprised, disgusted, neutral) using intensity scores of 100\%~\cite{williams_2023}. Stimuli were gender balanced (17 male, 18 female identities) and presented in randomized order. These trials were embedded within a broader experimental session that included additional tasks (not reported here).

\paragraph{Procedure.} 
Following HMD and EMG sensor placement, participants performed a facial expression re-enactment task. To elicit naturalistic expressions, we employed an emotion-focused instruction set: participants were asked to first internalize the displayed emotion before expressing it facially.
Each trial followed a fixed structure: (1) emotion label presentation (2\,s), during which participants prepared to experience the upcoming emotion; (2) facial expression image presentation (6\,s), during which participants expressed and maintained the emotion on the image on their own face; and (3) rest period displaying a fixation cross (20\,s ± 4\,s jitter), during which participants returned to a relaxed, neutral state. Participants were instructed to begin internalizing the emotion during the label presentation phase so that they could express it more naturally during the image presentation phase.  

\paragraph{Synchronization.} 
All EMG and video streams were synchronized using Lab Streaming Layer (LSL) (version:v1.16.4)~\cite{kothe2025lab}. The three GigE cameras were triggered by a common host, and VR event markers were inserted into the stream at trial onset.

\begin{figure}[t]
\centering
\includegraphics[width=\linewidth]{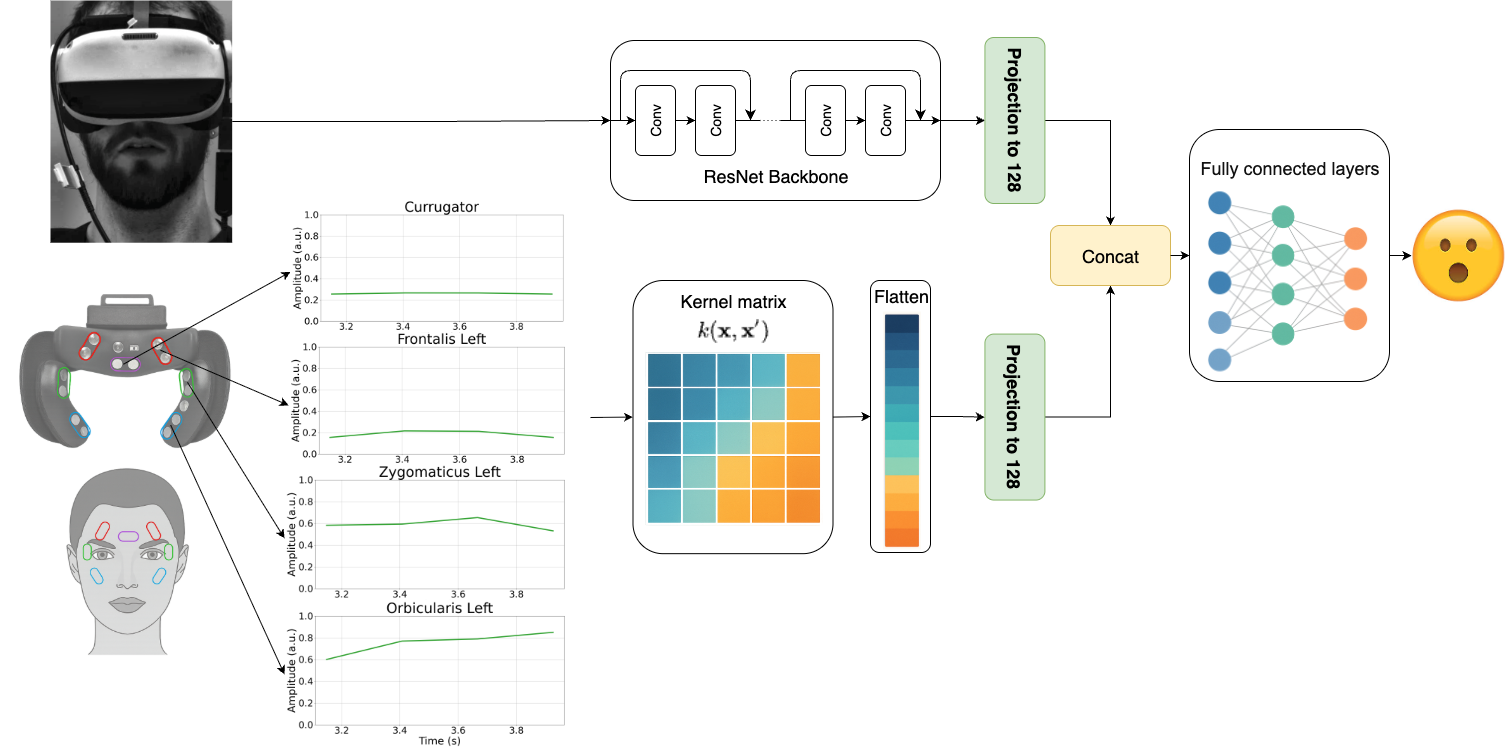}

  \caption{Overview of the proposed multimodal architecture for affect recognition in VR.  
The image branch (top) processes lower-face video frames using a ResNet backbone, projecting features into a 128-D embedding.  
The EMG branch (bottom) encodes signals from seven facial muscles into a RBF kernel matrix, which is then flattened and projected into a 128-D space.  
Both embeddings are concatenated and classified through fully connected layers into one of seven emotion categories.  
The modular design allows for the exchange of individual components: image backbones can be replaced by more modern encoding architectures (e.g., MobileNetv4~\cite{qin2024mobilenetv4}, Vision Transformer~\cite{dosovitskiy2020image}, Vision Mamba~\cite{Lianghui24_ViM}); the EMG branch can utilize handcrafted features or direct post-processed signals; and additional sensor modalities can be incorporated. Similarly, the classifier head can be adapted to experiment with different fusion strategies or architectures.}

\label{fig:net_diagram}
\end{figure}


\section{Method: Multimodal Two-Stream Classification}
\label{sec:method}

We propose a late-fusion architecture that integrates (i) a lower-face \emph{video stream} and (ii) an upper-face \emph{EMG stream} to classify seven emotions under HMD occlusion (Fig.~\ref{fig:net_diagram}). The design emphasizes modularity—either stream can be swapped or upgraded without changing the fusion contract.

\subsection{EMG Stream: Sensing and Preprocessing}
\label{subsec:emg_preproc}
Signals from seven facial muscles (Tab.~\ref{tab:au_mapping}) are processed using identical parameters across participants:
\begin{enumerate}
    \item \textbf{Filtering:} zero-phase IIR notch filter at 50\,Hz and its harmonics (up to 400\,Hz), followed by a 100--400\,Hz band-pass filter.

    \item \textbf{Epoching:} $[-1,6]$\,s relative to stimulus onset, with baseline subtraction (averaged signal amplitude from $-1$ to $0$\,s).
    \item \textbf{Amplitude clipping:} global clip per muscle at the maximum observed across participant-averaged trials to limit transient spikes.
    \item \textbf{Rectification:} full-wave rectification and 25\,ms Root Mean Square (RMS).
    \item \textbf{Temporal binning:} non-overlapping 250\,ms segments, following the approach of~\cite{PapervonLou}.
    \item \textbf{Normalization:} Min–max scaling to $[0,1]$ performed across all trials collectively (i.e., using the global minimum and maximum across all subjects), rather than separately for each subject.

\end{enumerate}

\subsection{Feature Extraction (EMG)}
Analysis used $[1,6]$\,s post-stimulus. Each trial was segmented with 1.0\,s windows and 0.25\,s hop. We used an RBF kernel representation as a compact, interpretable approximation of inter-muscle correlation structure, offering stability with limited data and low computational cost compared to deep sequence encoders. For each window, we concatenated the per-muscle scaled RMS values and computed the kernel matrix:
\begin{equation}
k(x_i, x_j) = \exp\!\left(- \frac{\lVert x_i - x_j \rVert^2}{2l^2} \right)
\end{equation}
The resulting matrix is a positive semi-definite (PSD) symmetric matrix that captures non-linear correlations between the seven EMG sensors within a 1\,s window. This matrix is flattened into a 784-D vector representation.  

For comparison, we also evaluated direct post-processed EMG features (28-D), where each dimension corresponds to one RMS value per channel and time bin, without kernelization.

\subsection{Video Stream: Lower-Face Features}
\label{subsec:video_features}
For each EMG window, we take the temporally aligned final frame from each of the three cameras (Sec.~\ref{Dataset}).This pairing of three lower-face images with a single EMG window resulted in 35649 multimodal samples across the dataset.
Following frame extraction, facial landmarks were localized using YOLO-Pose~\cite{YOLOPose, yolo11_ultralytics} localizes landmarks; we center a $480{\times}640$ crop at the nose (Fig.~\ref{fig:raw_yolo_examples}, bottom). Frames are normalized to $[0,1]$ and passed to a ResNet backbone initialized on the Extended Cohn-Kanade Dataset CK+~\cite{Cohn2010ck+}, providing a domain-relevant prior for facial expression recognition. Subsequently, channel-wise normalization is applied using dataset-specific statistics, with mean $\mu = [0.2948]$ and standard deviation $\sigma = [0.3002]$.

\begin{figure}[t]
    \centering
    \begin{subfigure}[t]{0.23\linewidth}
        \centering
        \includegraphics[width=\linewidth]{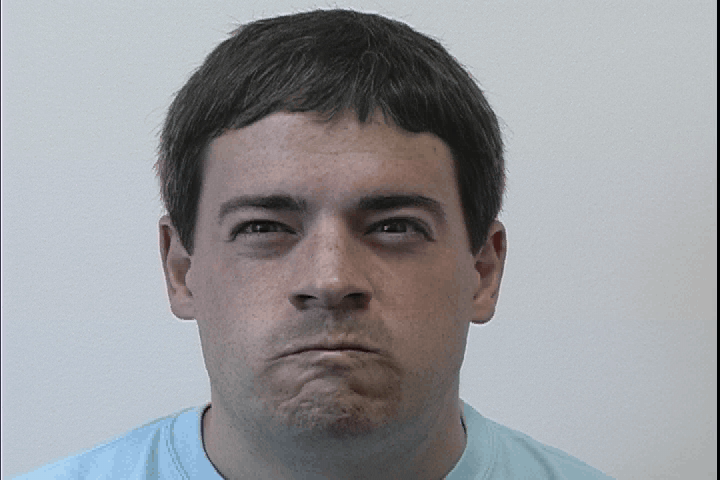}
        \caption{Original image}
        \label{fig:ck_a}
    \end{subfigure}
    \hfill
    \begin{subfigure}[t]{0.23\linewidth}
        \centering
        \includegraphics[width=\linewidth]{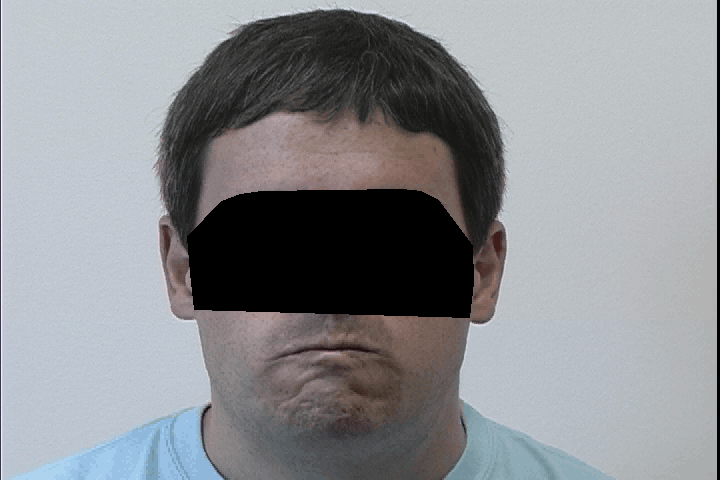}
        \caption{Black patch}
        \label{fig:ck_b}
    \end{subfigure}
    \hfill
    \begin{subfigure}[t]{0.23\linewidth}
        \centering
        \includegraphics[width=\linewidth]{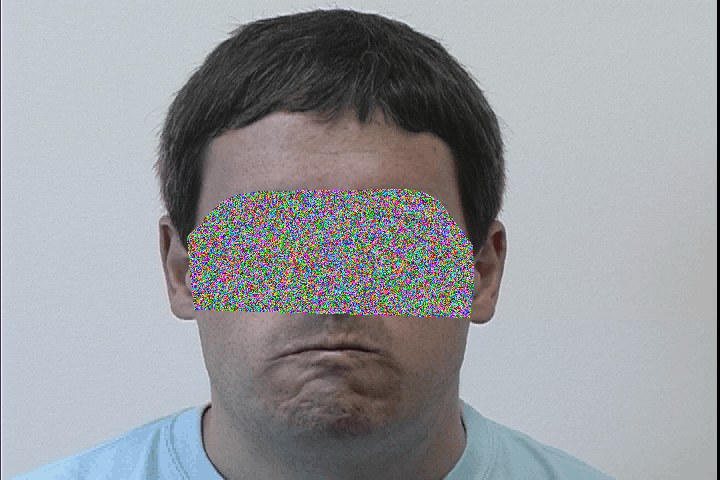}
        \caption{Random noise patch}
        \label{fig:ck_c}
    \end{subfigure}
    \hfill
    \begin{subfigure}[t]{0.23\linewidth}
        \centering
        \includegraphics[width=\linewidth]{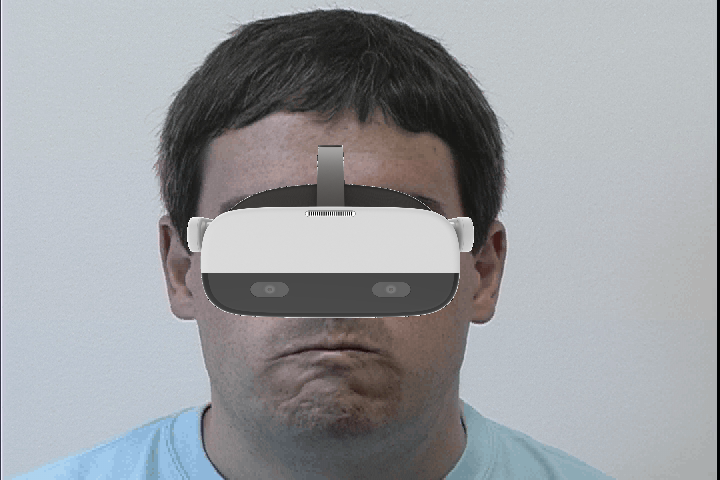}
        \caption{VR headset overlay}
        \label{fig:ck_d}
    \end{subfigure}

    \caption{Examples of CK+ occlusion augmentation.}
    \label{fig:ckplus_augmentation}
\end{figure}

To further adapt the pre-trained model to our occlusion setting, we augmented the CK+ training images by modifying the upper-face region. Each original image was supplemented with three augmented variants (Fig.~\ref{fig:ckplus_augmentation}):  
(i) black occlusion patches covering the eyes,  
(ii) random noise patches applied to the same region, and  
(iii) synthetic overlays of a VR headset.  
These augmentations encouraged the model to focus on lower-face cues while remaining robust to varying types of occlusion.

\subsection{Fusion Head and Embedding Spaces}
\label{subsec:fusion}
The image branch (ResNet-18/50 truncated before its classifier) and the EMG branch (either 784-D K-EMG or 28-D direct EMG) are projected into 128-dimensional embedding spaces.
We adopt a 128-dimensional projection head following the image encoder (as well as the EMG encoder), consistent with prior contrastive learning approaches~\cite{chen2020SimCLR,khosla2020SupContrast}. Given the size of the training set, this dimensionality provides a suitable trade-off between representational capacity, computational efficiency, and regularization; larger projection dimensions were therefore avoided to reduce the risk of overfitting.
The resulting embeddings are concatenated and fed to a fully connected fusion head consisting of two hidden layers with 256 and 128 units, respectively, interleaved with Batch Normalization and ReLU activations, and a final linear layer producing seven emotion logits(Fig.~\ref{fig:net_diagram}). This late-fusion design enables independent improvements to each modality or the fusion head (e.g., alternative vision encoders, handcrafted EMG features, or different fusion operators).

\subsection{Training and Evaluation} 
Models were trained using cross-entropy loss with Adam ($\text{lr}=1\mathrm{e}{-3}$, batch size 8) for 30 epochs. The selection of the model for testing was based on the macro-F1 validation.  
Given compute constraints (35k multimodal samples across 20 identities), we fixed a subject-disjoint test set of two identities and pooled the remaining 18 identities for training/validation. 
Validation batches were sampled at random (fixed seed 42) from the 18 training identities and used for early stopping and model selection. 
In preliminary experiments, 5-fold identity splits (16 train / 2 val / 2 test) did not improve test performance compared to training on the full set of 18 identities, while incurring substantially higher training time; the larger training pool exposed the model to more diverse EMG patterns, which we found beneficial in practice.

\section{Results}
We first conduct an \emph{ablation-driven model selection} to identify effective design choices for each stream and the fusion head. We then report final test results (shown in Table~\ref{tab:model_comparison}) on the held-out identities using the best configuration emerging from these ablations.

\paragraph{Ablation-Driven Model Selection Factors}
We query: (i) \textbf{modality} (Img, EMG, Img+EMG); (ii) \textbf{backbone} (ResNet-18, ResNet-50); (iii) \textbf{preprocessing} (with/without CLAHE\footnote{CLAHE is a method to adaptively improves local contrast by redistributing pixel intensities within small contextual regions while limiting noise amplification, thereby highlighting facial features more effectively. We apply CLAHE for each frame.}~\cite{CLAHE}); (iv) \textbf{EMG representation} (K-EMG 784-D vs.\ direct EMG 28-D).

\begin{table*}[t]
\centering
\caption{Testset performance comparison of models under different preprocessing and modality settings. All values represent means across the two held-out test subjects. All ResNet models were initialized with CK+ pretraining before fine-tuning on our dataset. 
OpenFace and SVM baselines do not rely on CK+ pretraining.
\textbf{Modality codes:} Img = images; K-EMG = kernel EMG (784-D); EMG = post-processed EMG (28-D).}
\begin{tabular}{lccc|cccc}
\toprule
\textbf{Model} & \textbf{Finetuned (ours)} & \textbf{CLAHE} & \textbf{Modality} & \textbf{Accuracy} & \textbf{F1} & \textbf{Precision} & \textbf{Recall} \\
\midrule
OpenFace 3 & --  & $\checkmark$ & Img & 0.16 & 0.10 & 0.11 & 0.14 \\
OpenFace 3 & --  & $\times$ & Img & 0.15 & 0.08 & 0.10 & 0.13 \\

ResNet-18 & $\times$ & $\checkmark$ & Img & 0.13 & 0.082 & 0.060 & 0.142 \\
ResNet-50 & $\times$ & $\checkmark$ & Img & 0.14 & 0.081 & 0.060 & 0.141 \\
ResNet-18 & $\times$ & $\times$ & Img & 0.14 & 0.083 & 0.060 & 0.143 \\
ResNet-50 & $\times$ & $\times$ & Img & 0.14 & 0.082 & 0.060 & 0.143 \\

\midrule

ResNet-18 & $\checkmark$ & $\checkmark$ & Img & 0.40 & 0.40 & 0.43 & 0.4 \\
ResNet-18  & $\checkmark$ & $\checkmark$ & Img+K-EMG & 0.41 & 0.42 & 0.51 & 0.41 \\
ResNet-50  & $\checkmark$ & $\checkmark$ & Img & 0.40 & 0.41 & 0.46 & 0.4 \\
ResNet-50$\dagger$  & $\checkmark$ & $\checkmark$ & Img+K-EMG & 0.50 & 0.51 & 0.55 & 0.50 \\

ResNet-18  & $\checkmark$ & $\times$ & Img & 0.32 & 0.30 & 0.33 & 0.32 \\
ResNet-18  & $\checkmark$ & $\times$ & Img+K-EMG & 0.39 & 0.35 & 0.38 & 0.39 \\
ResNet-50  & $\checkmark$ & $\times$ & Img & 0.39 & 0.39 & 0.44 & 0.39 \\
ResNet-50  & $\checkmark$ & $\times$ & Img+K-EMG & 0.40 & 0.40 & 0.43 & 0.40 \\
SVM  & $\checkmark$ & -- & K-EMG & 0.43 & 0.38 & 0.44 & 0.43 \\
\midrule
ResNet-18 & $\checkmark$ & $\checkmark$ & Img+EMG & 0.30 & 0.28 & 0.29 & 0.30 \\
ResNet-50 & $\checkmark$ & $\checkmark$ & Img+EMG & 0.38 & 0.36 & 0.40 & 0.38 \\
ResNet-18 & $\checkmark$ & $\times$ & Img+EMG & 0.35 & 0.28 & 0.31 & 0.35 \\
ResNet-50 & $\checkmark$ & $\times$ & Img+EMG & 0.34 & 0.29 & 0.36 & 0.34 \\
SVM  & $\checkmark$ & -- & EMG & 0.46 & 0.43 & 0.44 & 0.45 \\

\bottomrule
\end{tabular}
\label{tab:model_comparison}
\begin{tablenotes}
\footnotesize
\item $\dagger$ Inter-subject variability for this configuration: 
1st test subject achieved F1=0.66, while the 2nd test subject achieved F1=0.36, indicating substantial individual differences 
in cross-subject generalization.
\end{tablenotes}

\end{table*}


\begin{table*}[t]
\centering
\caption{Training and validation performance of ResNet-18 and ResNet-50 models under different modality settings.  
Metrics are reported after fine-tuning on our dataset, with and without CLAHE preprocessing.  
Results highlight that multimodal fusion (Img+EMG or Img+K-EMG) achieves substantially higher validation accuracy and F1 compared to image-only baselines, though in some cases validation scores approach ceiling levels, suggesting potential overfitting relative to held-out test results.}

\begin{tabular}{lcccccc}
\toprule
\textbf{Model} & \textbf{CLAHE} & \textbf{Modality} & \textbf{Train Acc.} & \textbf{Train F1} & \textbf{Val. Acc.} & \textbf{Val F1} \\
\midrule

ResNet-18 & $\checkmark$ & Img &  0.775 & 0.774 & 0.730 & 0.717 \\
ResNet-18 & $\times$ & Img & 0.770 & 0.770 & 0.753 & 0.743 \\
ResNet-50 & $\checkmark$ & Img &  0.850 & 0.850 & 0.868 & 0.865 \\
ResNet-50 & $\times$ & Img & 0.780 & 0.780 & 0.792 & 0.790 \\
\midrule
ResNet-18 & $\checkmark$ & Img+K-EMG & 0.908 & 0.908  & 0.950 & 0.950 \\
ResNet-18 & $\times$ & Img+K-EMG & 0.783 & 0.783 & 0.721 & 0.703 \\
ResNet-50 & $\checkmark$ & Img+K-EMG & 0.929 & 0.930 & 0.946 & 0.946 \\
ResNet-50 & $\times$ & Img+K-EMG & 0.788 & 0.788 & 0.818 & 0.817 \\
\midrule
ResNet-18 & $\checkmark$ & Img+EMG & 0.756 & 0.756 & 0.887 & 0.888 \\
ResNet-18 & $\times$ & Img+EMG & 0.776 & 0.775 & 0.895 & 0.896 \\
ResNet-50 & $\checkmark$ & Img+EMG & 0.844 & 0.844 & 0.938 & 0.934 \\
ResNet-50 & $\times$ & Img+EMG & 0.879 & 0.879 & 0.952 & 0.952 \\
\bottomrule
\end{tabular}
\label{tab:validation_metrics}
\end{table*}

\paragraph{Baseline performance.} 
Conventional image-only methods performed poorly under HMD occlusion. OpenFace-3.0~\cite{hu2025openface} yielded near-chance accuracy ($\sim$16\%) with F1 below 0.1, demonstrating that established SOTA FER pipelines fail when upper-face features are unavailable. Similarly, Our ResNet+prediction head models pretrained on our occlusion-augmented CK+ data—but not fine-tuned on the target VR dataset—did not transfer effectively, confirming that domain-specific adaptation is essential.

\paragraph{Effect of pretraining without fine-tuning.} 
To assess the value of domain adaptation, we evaluated ResNet-18 and ResNet-50 pretrained on CK+ (with occlusion augmentation) but not finetuned on our dataset. Performance dropped to chance level (F1 $\approx$0.08), underscoring that finetuning on domain-specific VR data is essential. This also validates our augmentation strategy: without adaptation, even occlusion-aware pretraining fails to transfer effectively.

\paragraph{Effect of backbone and preprocessing.} 
With finetuning, both ResNet-18 and ResNet-50 achieved comparable performance, with ResNet-50 slightly edging over using Image-only modality, Image+(Kernel EMG or post-processed EMG). CLAHE consistently improved results by up to 10 F1 points, especially benefiting subtle contrasts in the lower-face region. Without CLAHE, performance degraded, particularly for ResNet-18.

\paragraph{Role of EMG.} 
Unimodal EMG models again proved highly informative. Using direct post-processed EMG features (28-D), an SVM achieved the best unimodal score (Accuracy = 46\%, F1 = 0.43), outperforming both image-only baselines and kernel-based EMG SVMs. This demonstrates that compact, temporally structured features are a viable representation, sometimes surpassing higher-dimensional kernel embeddings.

\paragraph{Fusion benefits.} 
Multimodal late fusion systematically improved over unimodal baselines, but the benefits varied by EMG representation. Kernel EMG yielded the strongest overall results, with ResNet-50 + K-EMG reaching the highest performance (Accuracy = 50\%, F1 = 0.51). In contrast, fusion with direct 28-D post-processed EMG provided smaller gains, peaking at F1 $\approx$ 0.36. These findings suggest that the kernelized representation more effectively captures inter-muscle correlations relevant for emotion discrimination, and the deep model is beneficial at extracting these representations. 

\paragraph{Inter-subject variability.} Although fusion improved average performance, substantial variability emerged between the two test subjects. For the best-performing model (ResNet-50 + K-EMG with CLAHE), F1 scores ranged from 0.36 to 0.66. Without CLAHE, both subjects showed reduced but similarly variable performance (F1: 0.34–0.46). This dispersion underscores the challenge of cross-subject generalization and motivates future work with larger participant pools.

\begin{table}[ht!]
\centering
\caption{Illustrative per-class test accuracy for using image modality with ResNet-50 vs Multimodal ResNet-50+K-EMG (Both with CK+ pretraining, CLAHE, and finetuned on our dataset), highlighting the contribution of upper-face EMG under HMD occlusion for the different classes. 
Only Action Units measurable by the EMG sensors (Table~\ref{tab:au_mapping}) are listed.}
\begin{tabular}{lccc}
\hline
\textbf{Emotion} & \textbf{Upper-face AUs} & \textbf{Img (ResNet-50)} & \textbf{ResNet 50 Img + K-EMG} \\
\hline
\textbf{Happiness} & AU6, AU12 & 0.64 & 0.75 \\
\textbf{Surprise} & AU1, AU2 & 0.65 & 0.84 \\
\textbf{Anger} & AU4, AU7 & 0.18 & 0.23 \\
\textbf{Fear} & AU1, AU4 & 0.48 & 0.49 \\
\textbf{Sadness} & AU1, AU4 & 0.17 & 0.49 \\
\textbf{Disgust} & AU7 & 0.16 & 0.21 \\
\textbf{Neutral} & -- & 0.48 & 0.51 \\
\hline
\textbf{Macro Avg.} & — & 0.40 & 0.51 \\
\hline
\end{tabular}
\label{tab:per_class_recall}
\end{table}

\paragraph{Training and validation behavior.} 
Table~\ref{tab:validation_metrics} provides training and validation metrics across settings.  
We observe that:
\begin{itemize}
    \item Image-only models converged stably but plateaued at moderate validation scores ($\sim$0.72--0.86 F1).
    \item Kernel-EMG fusion achieved very high validation performance ($\sim$0.95 F1), but test performance did not reach the same level, indicating possible overfitting.
    \item Direct EMG fusion showed strong validation generalization (Val. F1 up to 0.95 with ResNet-50, no CLAHE), but these gains did not consistently translate to the test set.
\end{itemize}
A notable gap emerged between validation and held-out test performance (Tab.~\ref{tab:validation_metrics} vs. Tab.~\ref{tab:model_comparison}).  
Kernel-based multimodal models achieved near-ceiling validation scores (Val.\ F1 $\approx$0.95), yet their test F1 plateaued around 0.50.  
Similarly, fusion with direct post-processed EMG reached validation F1 above 0.93 but did not yield consistent improvements on the test set.
Note that our validation split was sampled from the same 18 training identities (not disjoint by subject), which—together with the smaller validation set—likely inflates validation scores relative to the subject-disjoint test set. 
Thus, the observed Val–Test gap reflects both this optimistic validation protocol and the inherent difficulty of cross-subject generalization. 
Contributing factors include: 
(i) limited dataset size (35k samples across 20 participants), 
(ii) high intra-subject consistency that benefits validation but does not transfer to unseen identities, and 
(iii) the relatively low variability of lab-elicited expressions compared to spontaneous VR interactions.

These findings emphasize the need for larger-scale multimodal corpora and stronger regularization strategies (e.g., temporal modeling, dropout, or domain adaptation) to bridge the gap between in-sample validation and cross-subject generalization in immersive VR.

\begin{figure}[t] 
\centering 

\includegraphics[width=0.49\linewidth]{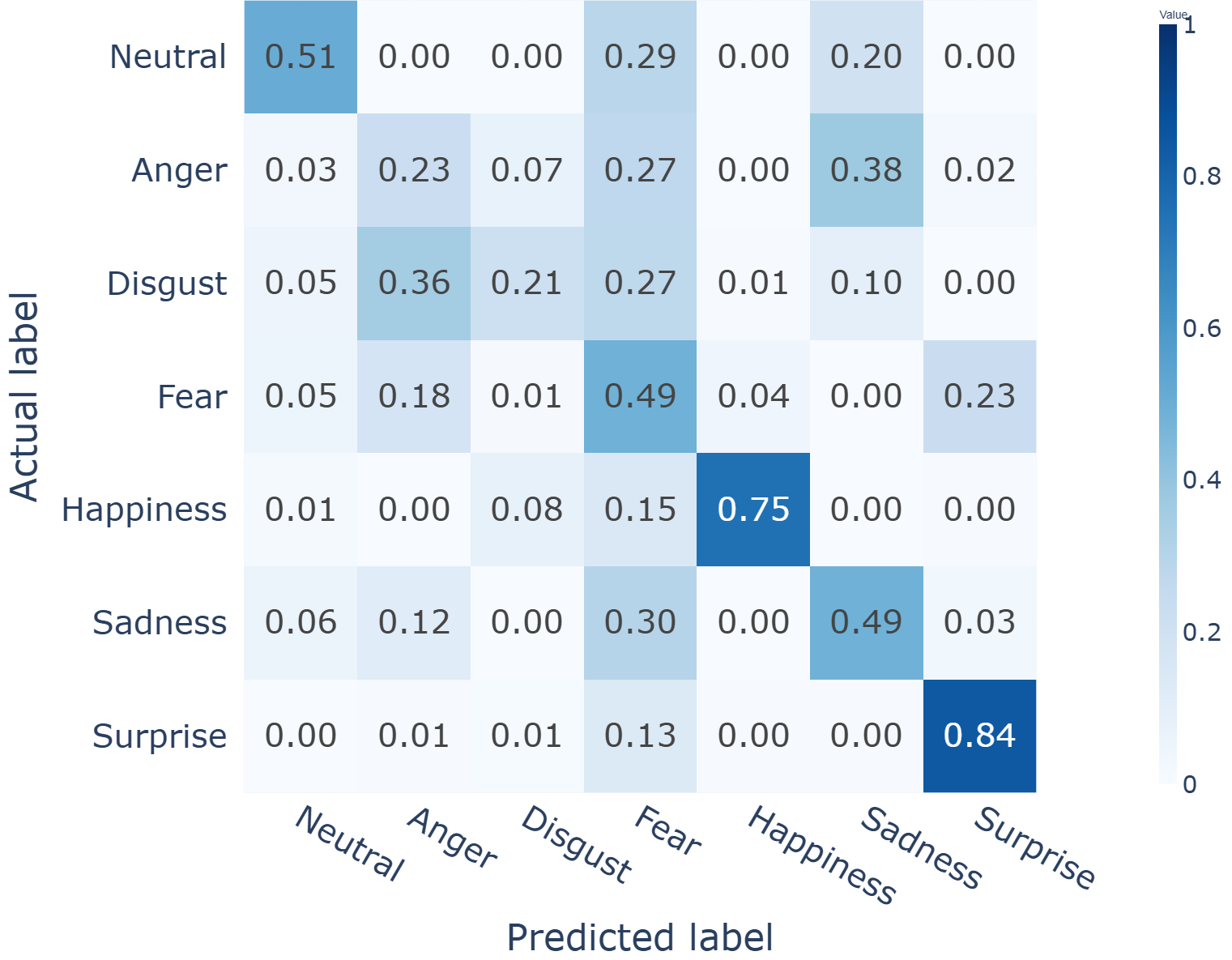} \includegraphics[width=0.49\linewidth]{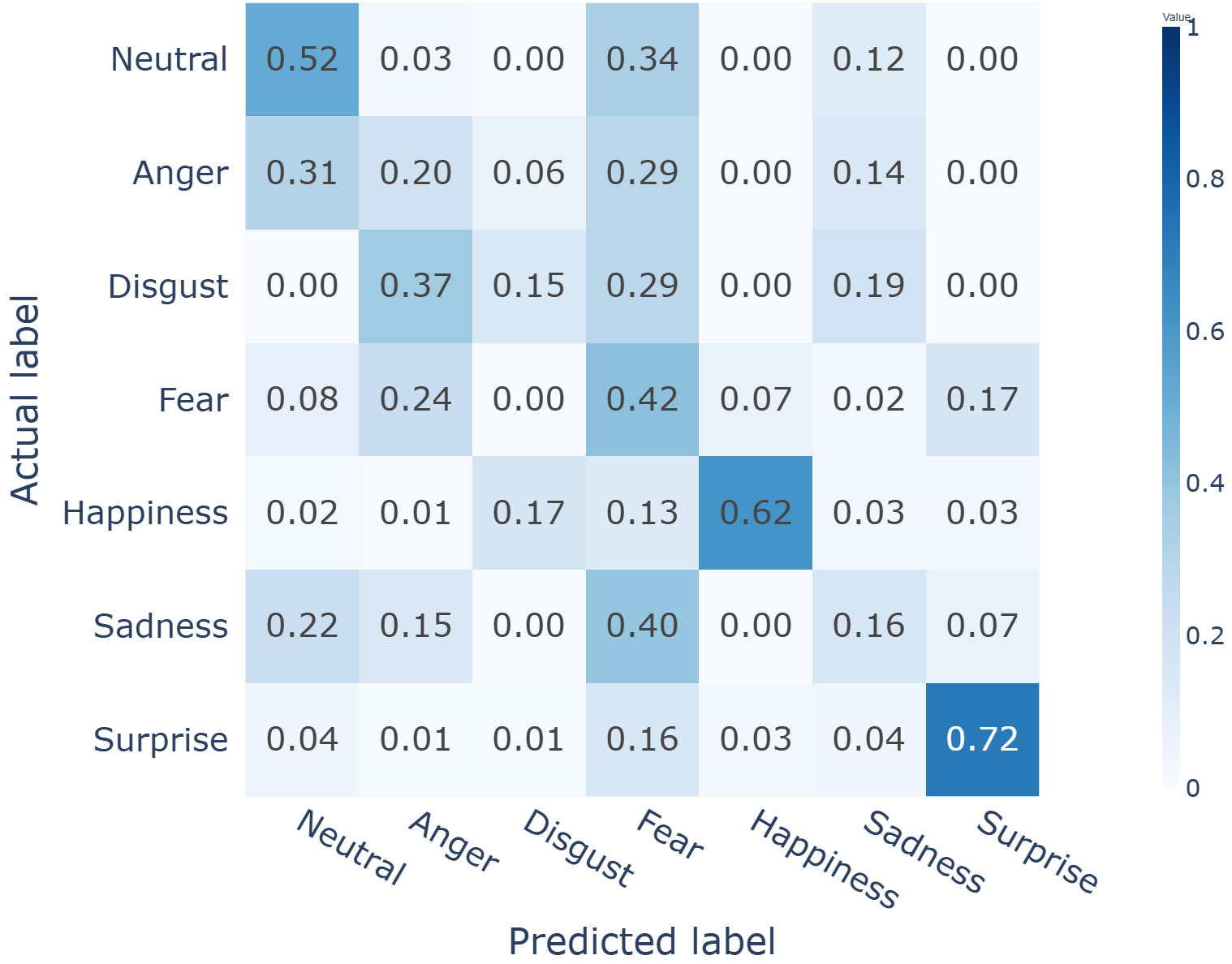}

\caption{Normalized confusion matrices for the multimodal ResNet-50 model with CLAHE preprocessing (left) and without CLAHE (right).  
Both models reliably recognize highly expressive emotions such as Surprise and Happiness, but CLAHE reduces confusions between Neutral and Sadness and improves recall for subtle categories.  
Values are row-normalized to indicate per-class recall.}
\label{fig:confmat_comp}
\end{figure}

\paragraph{Per-class analysis.} 
Table~\ref{tab:per_class_recall} reports per-class accuracy for image-only and multimodal models. Multimodal fusion with upper-face EMG yields consistent gains for emotions relying on occluded upper-face Action Units, with the largest improvements observed for Surprise (AU1/AU2: +19 points) and Sadness (AU1/AU4: +32 points). In contrast, improvements for mouth-dominant or neutral expressions are smaller, indicating that EMG primarily compensates for missing upper-face visual information under HMD occlusion rather than uniformly boosting performance across all classes.

\paragraph{Confusion matrix analysis.} 
Fig.~\ref{fig:confmat_comp} presents normalized confusion matrices for the multimodal ResNet-50 with and without CLAHE, respectively.  
Surprise and Happiness were recognized most reliably, reaching up to 0.84 and 0.75 true positive rates with CLAHE, while Neutral was also consistently well-identified ($\sim$0.5 recall).  
In contrast, Anger and Disgust were frequently confused with Fear and Sadness, suggesting overlapping muscle activation patterns in the upper face when occluded by the HMD.  

Notably, the strong performance for Happiness is consistent with prior findings in human emotion perception, which show that happy expressions are detected more rapidly and robustly than other emotions due to their high visual salience, particularly driven by mouth-related features~\cite{calvo2008detection}. This aligns with our occlusion setting, where the lower face remains visible and informative, whereas emotions that rely more heavily on subtle upper-face cues (e.g., Anger, Disgust) remain challenging to disambiguate.  

Finally, CLAHE enhanced discriminability across most classes—especially for Happiness and Surprise—while without CLAHE the model exhibited increased confusions between Sadness and Neutral. Overall, the fusion model captures strong signals for highly salient expressions (Happiness, Surprise) but remains challenged by subtler emotions (Anger, Disgust).

\paragraph{Summary.} 
Overall, our results demonstrate (i) the limited utility of image-only facial emotion recognition under HMD occlusion, even with occlusion-aware pretraining; (ii) the discriminative strength of facial EMG as a complementary modality; (iii) the superiority of kernel-based facial EMG fusion over direct 28-D features for multimodal integration; and (iv) the importance of careful preprocessing and finetuning to achieve robust performance in immersive VR settings.

\section{Discussion}
\label{Conclusion}

We demonstrate the advantages of multimodal fusion for facial emotion recognition under HMD occlusion. Conventional computer vision pipelines such as OpenFace-3, which excel on full-face datasets, collapsed to near-chance levels when applied to lower-face-only inputs in VR. This underscores the limitations of directly transferring existing facial emotion recognition methods into immersive environments.

In contrast, facial EMG signals provided a robust unimodal alternative. Despite being restricted to seven upper-face channels, EMG alone outperformed vision-only baselines and captured discriminative muscle activation patterns associated with emotional expressions. This finding is consistent with prior evidence that surface EMG remains effective under occlusion~\cite{kiprijanovska2022facial}. Previous electroencephalography studies in VR confirm that even low-SNR biosignals remain reliable under HMD conditions~\cite{weber2021_vrinference}, further supporting the robustness of higher-SNR facial EMG in our setup. Interestingly, our ablations revealed that compact post-processed EMG (28-D RMS values) and higher-dimensional kernel embeddings (784-D) behave differently: the kernelized form provides stronger fusion benefits, while the direct representation proves more stable in unimodal settings.

The strongest contribution emerged from multimodal fusion. Combining lower-face images with EMG consistently improved recognition rates, particularly when using ResNet-50 backbones with CLAHE preprocessing. The best multimodal configuration yielded a 10-point macro-F1 gain over the corresponding image-only baseline. Confusion matrix analyses further revealed that highly expressive expressions such as Happiness and Surprise benefited the most from fusion. By contrast, subtler emotions such as Anger and Disgust remained difficult to disambiguate, often being misclassified as Fear or Sadness. These systematic confusions point to overlapping muscle activations and suggest the need for richer multimodal inputs or temporal context modeling.

From a methodological perspective, pretraining on CK+ was essential to stabilize training on our VR dataset. Without fine-tuning, pretrained networks performed at chance level, highlighting the importance of domain adaptation. Moreover, although validation metrics suggested near-ceiling performance for multimodal models (Val. F1 $\sim$0.95), generalization to unseen test identities was substantially weaker (Test F1 $\sim$0.50). This gap reflects overfitting to intra-subject consistencies and underscores the challenge of robust cross-subject affect recognition in VR.

Finally, our dataset contributes a novel resource for affective computing in immersive environments: roughly 35k paired image–EMG samples across seven emotions, collected from 20 participants under subject-exclusive splits. To our knowledge, this is the first dataset combining synchronized external video and upper-face EMG during VR exposure, and it enables systematic evaluation of multimodal facial emotion recognition under realistic occlusion.

\subsection*{Limitations}

\textbf{Evaluation protocol and generalization.} We adopted a pragmatic training scheme that prioritized compute efficiency, enabling us to run a broad set of ablations within fixed resources. Rather than full LOSO or K-fold identity cross-validation, we employed a fixed subject-disjoint test set (2 identities) and trained on the remaining 18 participants. This design maximized experimental coverage while preserving subject independence in testing. Preliminary trials with identity-fold splits did not yield higher test performance than training on the larger pool of 18 subjects, which provided the model with more diverse EMG signals in practice.

\textbf{Dataset scale and demographics.} Our dataset is relatively small (20 participants)  and reflects a limited demographic range. Subject-specific EMG patterns are known to vary substantially, and the limited number of identities constrains the diversity of muscle activation profiles seen during training. Larger and more demographically diverse datasets are needed to better quantify robustness across users and to reduce overfitting to participant-specific idiosyncrasies.
Our Dataset represents a critical first step toward tackling VR occlusion challenges. The controlled setting allowed us to tightly synchronize video and EMG, and to systematically probe multimodal fusion under HMD constraints. Importantly, the dataset design and open sharing agreement establish a foundation for future work to extend toward larger and broader demographics.

\textbf{Ecological validity of emotion elicitation.}
Our emotion induction relied on instructed imitation of static, high-intensity NimStim facial-expression images. This design provides strong experimental control and balanced classes, but it constrains ecological validity in two important ways: (i) the captured behavior is primarily \emph{posed} expression production rather than \emph{spontaneous} affective display, and (ii) static stimuli and a non-interactive trial structure do not reflect the dynamics of social VR training scenarios (e.g., conversational turn-taking, speech, head motion, cognitive load, and interactive agents). Consequently, the reported performance may not directly transfer to naturalistic VR interactions without additional validation.

\textbf{Limited temporal modeling.}
Although we apply temporal preprocessing (epoching, RMS extraction, and sliding windows), each window is classified independently and the model does not explicitly learn cross-window dynamics. This likely contributes to confusion among subtle emotions and to reduced cross-subject generalization. Incorporating lightweight temporal sequence models (e.g., LSTM/Mamba/Transformer over EMG and/or visual embeddings) and trial-level aggregation is a promising direction to improve robustness.

Finally, we observed a substantial gap between validation (Val.\ F1 $\approx$ 0.95) and subject-disjoint test performance (Test F1 $\approx$ 0.50). This reflects the combined effects of limited training identities, subject-specific variability in facial EMG patterns, and optimistic validation due to non-disjoint splits, rather than indicating failure of the multimodal fusion principle itself. These results highlight the inherent difficulty of cross-subject affect recognition in VR and motivate future work using larger cohorts, stricter identity-based evaluation protocols (e.g., LOSO), and explicit temporal modeling.

\subsection{Conclusion}
We introduced a multimodal framework for emotion recognition in VR that integrates lower-face video with upper-face EMG, directly addressing the occlusion challenge imposed by HMDs. Experiments with 20 participants demonstrated that multimodal fusion substantially outperforms unimodal baselines, with the greatest gains for salient emotions such as Happiness and Surprise.  

Our contributions are threefold: (i) release of a novel VR dataset with synchronized multimodal affective recordings, (ii) analysis of kernel-based versus direct EMG encodings, and (iii) demonstration of the benefits of pretraining and multimodal fusion under HMD occlusion.  

Although our focus here was methodological validation rather than deployment, the proposed architecture is computationally lightweight: ResNet backbones and compact EMG representations enable real-time inference on modern GPUs. This makes the approach directly suitable for affect-adaptive VR scenarios, such as communication training or therapy, which we aim to evaluate in future applied studies. Future work will also extend the dataset with additional physiological modalities (e.g., skin conductance, heart rate), incorporate temporal sequence modeling for more robust recognition of subtle emotions, and explore domain adaptation to reduce overfitting and improve cross-subject generalization. By bridging physiological and visual modalities, we aim to enable more robust, ecologically valid affect recognition in immersive environments.

\section{ACKNOWLEDGMENTS}
This research was funded by the German Ministry of Education and Research (K3VR, 13N16388),
the Max Planck Society, and the Fraunhofer Society (project NeuroHum).


\section*{ETHICAL IMPACT STATEMENT}

This research involves collecting and analysing biometric data, including facial expressions and EMG signals, which constitute sensitive personal information. Thus, all participants were informed about the study and its objective and provided written informed consent for participation and for the collection, analysis, and distribution of their physiological and camera data. All data were stored and analysed in pseudonymised form. The study was approved by the local ethics committee (Ethics Committee of Fraunhofer HHI, following the DFG´s Code of Conduct "Safeguarding Good Research Practice"~\cite{dfg_2025}). 

Beyond our study-specific safeguards, emotion recognition technologies raise significant privacy concerns as they can reveal intimate psychological states without explicit user awareness. To empower the user, consent and data management tools have been recommended~\cite{barker_2025} that proactively gather informed consent from the user and give transparent feedback when sensitive data is being recorded.

Emotion recognition systems can exhibit cultural, demographic, and individual biases that may lead to unfair or inaccurate interpretations of emotional states. Visualization dashboards with real-time insights into the data flow and the resulting decision-making may contribute to higher accuracy, trust, and unbiased performance~\cite{barker_2025}. 

The system lacks access to situational context, verbal content, or interaction history, all of which are essential for accurate affect interpretation. Therefore, outputs should be treated as probabilistic indicators requiring human judgment, not definitive emotional labels.

\paragraph{Regulatory Considerations.}
Emotion recognition systems based on biometric signals may fall within the scope of high-risk AI under the EU AI Act (Regulation 2024/1689), particularly when deployed in employment contexts (e.g., communication training for professionals) or healthcare settings (e.g., therapeutic 
interventions). Such systems may require conformity assessments, human oversight mechanisms, and transparency obligations. We emphasize that the current work constitutes a prototype intended for controlled experimental settings, not a deployment-ready product. Any translation to applied contexts would require (i) compliance with applicable AI regulations, including mandated risk assessments and conformity procedures; (ii) ethical governance through stakeholder engagement and continuous monitoring; and (iii) where research is involved, institutional review board approval for studies involving human participants.

\subsection*{Dataset Availability and Risk-Mitigation}
\label{datasetAvailability}
We share access to the full dataset upon request; access requires acceptance of an ethical use and data-privacy-compliant agreement. 
The release includes synchronized lower-face video, facial EMG signals, and metadata, together with a reference split (18 training/validation identities, 2 test identities) as a starting point. 
Researchers are encouraged to adopt alternative evaluation protocols (e.g., LOSO, 5-fold identity CV) using the provided subject IDs and metadata; scripts for reproducing our reference split are included. 
To mitigate risks of misuse, the dataset is intended solely for non-commercial research on affective computing and VR interaction. Redistribution or application for surveillance, profiling, or covert monitoring is explicitly prohibited under the usage agreement.

\section*{Declaration on Generative AI}
 During the preparation of this work, the authors used different commercial LLMs in order to: Grammar and spelling check. After using these tools, the authors reviewed and edited the content as needed and take full responsibility for the publication’s content. 

\bibliography{manuscript-citation}

@String{Computing = "Computing" }

@String{Computer = "{IEEE} Computer" }

@String{Academic = "Academic Press" }

@InProceedings{chen2020SimCLR,
  title = 	 {A Simple Framework for Contrastive Learning of Visual Representations},
  author =       {Chen, Ting and Kornblith, Simon and Norouzi, Mohammad and Hinton, Geoffrey},
  booktitle = 	 {Proceedings of the 37th International Conference on Machine Learning},
  pages = 	 {1597--1607},
  year = 	 {2020},
  editor = 	 {III, Hal Daumé and Singh, Aarti},
  volume = 	 {119},
  series = 	 {Proceedings of Machine Learning Research},
  month = 	 {13--18 Jul},
  publisher =    {PMLR}
}

@inproceedings{khosla2020SupContrast,
author = {Khosla, Prannay and Teterwak, Piotr and Wang, Chen and Sarna, Aaron and Tian, Yonglong and Isola, Phillip and Maschinot, Aaron and Liu, Ce and Krishnan, Dilip},
title = {Supervised contrastive learning},
year = {2020},
isbn = {9781713829546},
publisher = {Curran Associates Inc.},
address = {Red Hook, NY, USA},
booktitle = {Proceedings of the 34th International Conference on Neural Information Processing Systems},
articleno = {1567},
numpages = {13},
location = {Vancouver, BC, Canada},
series = {NIPS '20}
}

@article{calvo2008detection, title={Detection of emotional faces: salient physical features guide effective visual search.}, author={Calvo, Manuel G and Nummenmaa, Lauri}, journal={Journal of Experimental Psychology: General}, volume={137}, number={3}, pages={471}, year={2008}, publisher={American Psychological Association} }

@INPROCEEDINGS {Ortmann2024EmojiHeroVR,
author = { Ortmann, Thorben and Wang, Qi and Putzar, Larissa },
booktitle = { 2024 12th International Conference on Affective Computing and Intelligent Interaction (ACII) },
title = {{ EmojiHeroVR: A Study on Facial Expression Recognition Under Partial Occlusion from Head-Mounted Displays }},
year = {2024},
volume = {},
ISSN = {},
pages = {80-88},
doi = {10.1109/ACII63134.2024.00014},
url = {https://doi.ieeecomputersociety.org/10.1109/ACII63134.2024.00014},
publisher = {IEEE Computer Society},
address = {Los Alamitos, CA, USA},
month =sep}

@ARTICLE{Chanel2011SMC,
  author={Chanel, Guillaume and Rebetez, Cyril and Bétrancourt, Mireille and Pun, Thierry},
  journal={IEEE Trans. Syst. Man Cybern. A:Syst. Hum.}, 
  title={Emotion Assessment From Physiological Signals for Adaptation of Game Difficulty}, 
  year={2011},
  volume={41},
  number={6},
  pages={1052-1063},
  doi={10.1109/TSMCA.2011.2116000}
}

@article{MarinMorales2020Sensors,
AUTHOR = {Marín-Morales, Javier and Llinares, Carmen and Guixeres, Jaime and Alcañiz, Mariano},
TITLE = {Emotion Recognition in Immersive Virtual Reality: From Statistics to Affective Computing},
JOURNAL = {Sensors},
VOLUME = {20},
YEAR = {2020},
NUMBER = {18},
ARTICLE-NUMBER = {5163},
PubMedID = {32927722},
ISSN = {1424-8220},
DOI = {10.3390/s20185163}
}

@article{Gnacek2024SciData,
 title={AVDOS-VR: Affective Video Database with Physiological Signals and Continuous Ratings Collected Remotely in VR}, volume={11}, DOI={10.1038/s41597-024-02953-6}, number={1}, journal={Scientific Data}, author={Gnacek, Michal and Quintero, Luis and Mavridou, Ifigeneia and Balaguer-Ballester, Emili and Kostoulas, Theodoros and Nduka, Charles and Seiss, Ellen}, year={2024}, month=jan, pages={132}, language={en} }

@article{Polo2025CBM,
title = {Advancing emotion recognition with Virtual Reality: A multimodal approach using physiological signals and machine learning},
journal = {Computers in Biology and Medicine},
volume = {193},
pages = {110310},
year = {2025},
issn = {0010-4825},
doi = {https://doi.org/10.1016/j.compbiomed.2025.110310},
author = {Edoardo Maria Polo and Francesco Iacomi and Alberto Valdes Rey and Davide Ferraris and Alessia Paglialonga and Riccardo Barbieri}
}

@inproceedings{Hickson2019WACV,
author={Hickson, Steven and Dufour, Nick and Sud, Avneesh and Kwatra, Vivek and Essa, Irfan},
  booktitle={Proc. IEEE WACV}, 
  title={Eyemotion: Classifying Facial Expressions in VR Using Eye-Tracking Cameras}, 
  year={2019},
  volume={},
  number={},
  pages={1626-1635},
  doi={10.1109/WACV.2019.00178}}

@inproceedings{Murakami2019SIGGRAPH,
author = {Murakami, Masaaki and Kikui, Kosuke and Suzuki, Katsuhiro and Nakamura, Fumihiko and Fukuoka, Masaaki and Masai, Katsutoshi and Sugiura, Yuta and Sugimoto, Maki},
title = {AffectiveHMD: facial expression recognition in head mounted display using embedded photo reflective sensors},
year = {2019},
isbn = {9781450363082},
publisher = {Association for Computing Machinery},
address = {New York, NY, USA},
doi = {10.1145/3305367.3335039},
booktitle = {ACM SIGGRAPH 2019 Emerging Technologies},
articleno = {7},
numpages = {2},
location = {Los Angeles, California},
series = {SIGGRAPH '19}
}

@article{Gjoreski2022SciRep,
title={Facial EMG sensing for monitoring affect using a wearable device}, volume={12}, DOI={10.1038/s41598-022-21456-1}, number={1}, journal={Scientific Reports}, author={Gjoreski, Martin and Kiprijanovska, Ivana and Stankoski, Simon and Mavridou, Ifigeneia and Broulidakis, M. John and Gjoreski, Hristijan and Nduka, Charles}, year={2022}, month=oct, pages={16876} }

@inproceedings{Gjoreski2021Ubicomp,
author = {Gjoreski, Hristijan and I. Mavridou, Ifigeneia and Fatoorechi, Mohsen and Kiprijanovska, Ivana and Gjoreski, Martin and Cox, Graeme and Nduka, Charles},
title = {emteqPRO: Face-mounted Mask for Emotion Recognition and Affective Computing},
year = {2021},
isbn = {9781450384612},
publisher = {Association for Computing Machinery},
address = {New York, NY, USA},
doi = {10.1145/3460418.3479276},
booktitle = {Adjunct Proceedings of the 2021 ACM International Joint Conference on Pervasive and Ubiquitous Computing and Proceedings of the 2021 ACM International Symposium on Wearable Computers},
pages = {23–25},
numpages = {3},
location = {Virtual, USA},
series = {UbiComp/ISWC '21 Adjunct}
}

@article{Khezri2015Fusion,
title = {Reliable emotion recognition system based on dynamic adaptive fusion of forehead biopotentials and physiological signals},
journal = {Computer Methods and Programs in Biomedicine},
volume = {122},
number = {2},
pages = {149-164},
year = {2015},
issn = {0169-2607},
doi = {https://doi.org/10.1016/j.cmpb.2015.07.006},
author = {Mahdi Khezri and Mohammad Firoozabadi and Ahmad Reza Sharafat}
}

@article{indrasiri2024multimodal,
      title={VR Based Emotion Recognition Using Deep Multimodal Fusion With Biosignals Across Multiple Anatomical Domains}, 
      author={Pubudu L. Indrasiri and Bipasha Kashyap and Chandima Kolambahewage and Bahareh Nakisa and Kiran Ijaz and Pubudu N. Pathirana},
      year={2024},
      eprint={2412.02283},
      archivePrefix={arXiv},
      primaryClass={eess.SP},
      url={https://arxiv.org/abs/2412.02283}, 
}

@article{NimStim_dataset,
author = {Tottenham, Nim and Tanaka, James and Leon, Andrew and Mccarry, Thomas and Nurse, Marcella and Hare, Todd and Marcus, David and Westerlund, Alissa and Casey, Bj and Nelson, Charles},
year = {2009},
month = {07},
pages = {242-9},
title = {The NimStim set of Facial Expressions: Judgments from Untrained Research Participants},
volume = {168},
journal = {Psychiatry research},
doi = {10.1016/j.psychres.2008.05.006}
}

@article{EmteqLabsPaper,
author = {Gnacek, Michal and Broulidakis, John and Mavridou, Ifigeneia and Fatoorechi, Mohsen and Seiss, Ellen and Kostoulas, Theodoros and Balaguer-Ballester, Emili and Kiprijanovska, Ivana and Rosten, Claire and Nduka, Charles},
year = {2022},
month = {03},
pages = {781218},
title = {EmteqPRO—Fully Integrated Biometric Sensing Array for Non-Invasive Biomedical Research in Virtual Reality},
volume = {3},
journal = {Frontiers in Virtual Reality},
doi = {10.3389/frvir.2022.781218}
}

@ARTICLE{PapervonLou,
  author={Lou, Jianwen and Wang, Yiming and Nduka, Charles and Hamedi, Mahyar and Mavridou, Ifigeneia and Wang, Fei-Yue and Yu, Hui},
  journal={IEEE Trans. Multimed.}, 
  title={Realistic Facial Expression Reconstruction for VR HMD Users}, 
  year={2020},
  volume={22},
  number={3},
  pages={730-743},
  doi={10.1109/TMM.2019.2933338}}

@article{diemer2015impact,
  title={The impact of perception and presence on emotional reactions: a review of research in virtual reality},
  author={Diemer, Julia and Alpers, Georg W and Peperkorn, Henrik M and Shiban, Youssef and M{\"u}hlberger, Andreas},
  journal={Frontiers in psychology},
  volume={6},
  pages={26},
  year={2015},
  publisher={Frontiers Media SA}
}

@article{SlaterWilbur1997FIVE,
    author = {Slater, Mel and Wilbur, Sylvia},
    title = {A Framework for Immersive Virtual Environments (FIVE): Speculations on the Role of Presence in Virtual Environments},
    journal = {Presence: Teleoperators and Virtual Environments},
    volume = {6},
    number = {6},
    pages = {603-616},
    year = {1997},
    month = {12},
    doi = {10.1162/pres.1997.6.6.603},
    url = {https://doi.org/10.1162/pres.1997.6.6.603},
    eprint = {https://direct.mit.edu/pvar/article-pdf/6/6/603/1623151/pres.1997.6.6.603.pdf},
}

@article{Felnhofer2015VRemotions,
title = "Is virtual reality emotionally arousing? Investigating five emotion inducing virtual park scenarios",
author = "A. Felnhofer and O. Kothgassner and Mareike Schmidt and Heinzle, {A.- K.} and Leon Beutl and Helmut Hlavacs and I. Kryspin-Exner",
note = "Publisher Copyright: {\textcopyright} 2015 Elsevier Ltd. All rights reserved.",
year = "2015",
month = oct,
day = "1",
doi = "10.1016/j.ijhcs.2015.05.004",
language = "English",
volume = "82",
pages = "48--56",
journal = "International Journal of Human-Computer Studies",
issn = "1071-5819",
publisher = "Elsevier",
}

@article{tian2021emotional,
  title={Emotional arousal in 2D versus 3D virtual reality environments},
  author={Tian, Feng and Hua, Minlei and Zhang, Wenrui and Li, Yingjie and Yang, Xiaoli},
  journal={PloS one},
  volume={16},
  number={9},
  pages={e0256211},
  year={2021},
  publisher={Public Library of Science San Francisco, CA USA}
}

@article{sanchez2005presence,
  title={From presence to consciousness through virtual reality},
  author={Sanchez-Vives, Maria V and Slater, Mel},
  journal={Nature reviews neuroscience},
  volume={6},
  number={4},
  pages={332--339},
  year={2005},
  publisher={Nature Publishing Group UK London}
}

@article{hofmann2021decoding,
  title={Decoding subjective emotional arousal from EEG during an immersive virtual reality experience},
  author={Hofmann, Simon M and Klotzsche, Felix and Mariola, Alberto and Nikulin, Vadim and Villringer, Arno and Gaebler, Michael},
  journal={elife},
  volume={10},
  pages={e64812},
  year={2021},
  publisher={eLife Sciences Publications Limited}
}

@article{marin2021heart,
  title={Heart rate variability analysis for the assessment of immersive emotional arousal using virtual reality: Comparing real and virtual scenarios},
  author={Mar{\'\i}n-Morales, Javier and Higuera-Trujillo, Juan Luis and Guixeres, Jaime and Llinares, Carmen and Alca{\~n}iz, Mariano and Valenza, Gaetano},
  journal={PloS one},
  volume={16},
  number={7},
  pages={e0254098},
  year={2021},
  publisher={Public Library of Science San Francisco, CA USA}
}

@article{McCall_2015_Physiophenomenology, title={Physiophenomenology in retrospect: Memory reliably reflects physiological arousal during a prior threatening experience}, volume={38}, url={https://www.sciencedirect.com/science/article/pii/S1053810015300386}, DOI={10.1016/j.concog.2015.09.011}, 
journal={Consciousness and Cognition}, author={McCall, Cade and Hildebrandt, Lea K. and Bornemann, Boris and Singer, Tania}, year={2015}, month=dec, pages={60–70} }

@article{Peperkorn_2014, title={Triggers of Fear: Perceptual Cues Versus Conceptual Information in Spider Phobia}, volume={70}, url={https://onlinelibrary.wiley.com/doi/abs/10.1002/jclp.22057}, DOI={10.1002/jclp.22057},  number={7}, journal={Journal of Clinical Psychology}, author={Peperkorn, Henrik M. and Alpers, Georg W. and Mühlberger, Andreas}, year={2014}, pages={704–714}, language={en} }

@article{maples2017treatment,
  title={The use of virtual reality technology in the treatment of anxiety and other psychiatric disorders},
  author={Maples-Keller, Jessica L and Bunnell, Brian E and Kim, Sae-Jin and Rothbaum, Barbara O},
  journal={Harvard review of psychiatry},
  volume={25},
  number={3},
  pages={103--113},
  year={2017},
  publisher={LWW}
}

@article{Ekman_1992, title={An argument for basic emotions}, volume={6}, url={https://doi.org/10.1080/02699939208411068}, DOI={10.1080/02699939208411068}, note={Publisher: Routledge}, number={3–4}, journal={Cognition and Emotion}, author={Ekman, Paul}, year={1992}, month=may, pages={169–200} }

@inproceedings{kiprijanovska2022facial,
  title={Facial Expression Recognition using Facial Mask with EMG Sensors.},
  author={Kiprijanovska, Ivana and Sazdov, Borjan and Majstoroski, Martin and Stankoski, Simon and Gjoreski, Martin and Nduka, Charles and Gjoreski, Hristijan},
  booktitle={VR4Health@ MUM},
  pages={23--28},
  year={2022}
}

@article{luma2022vreed,
author = {Tabbaa, Luma and Searle, Ryan and Bafti, Saber Mirzaee and Hossain, Md Moinul and Intarasisrisawat, Jittrapol and Glancy, Maxine and Ang, Chee Siang},
title = {VREED: Virtual Reality Emotion Recognition Dataset Using Eye Tracking \& Physiological Measures},
year = {2022},
issue_date = {Dec 2021},
publisher = {Association for Computing Machinery},
address = {New York, NY, USA},
volume = {5},
number = {4},
url = {https://doi.org/10.1145/3495002},
doi = {10.1145/3495002},
journal = {Proc. ACM Interact. Mob. Wearable Ubiquitous Technol.},
month = dec,
articleno = {178},
numpages = {20}
}

@ARTICLE{Lohesara2023HEADSET,
  author={Lohesara, Fatemeh Ghorbani and Freitas, Davi Rabbouni and Guillemot, Christine and Eguiazarian, Karen and Knorr, Sebastian},
  journal={IEEE Trans. Vis. Comput. Graph.}, 
  title={HEADSET: Human Emotion Awareness under Partial Occlusions Multimodal DataSET}, 
  year={2023},
  volume={29},
  number={11},
  pages={4686-4696},
  doi={10.1109/TVCG.2023.3320236}}

@article{Thies2018FaceVR,
author = {Thies, Justus and Zollh\"{o}fer, Michael and Stamminger, Marc and Theobalt, Christian and Nie\ss{}ner, Matthias},
title = {FaceVR: Real-Time Gaze-Aware Facial Reenactment in Virtual Reality},
year = {2018},
issue_date = {April 2018},
publisher = {Association for Computing Machinery},
address = {New York, NY, USA},
volume = {37},
number = {2},
issn = {0730-0301},
url = {https://doi.org/10.1145/3182644},
doi = {10.1145/3182644},
journal = {ACM Trans. Graph.},
month = jun,
articleno = {25},
numpages = {15}
}

@INPROCEEDINGS{Numan2021HMDRemoval,
  author={Numan, Nels and Haar, Frank ter and Cesar, Pablo},
  booktitle={Proc. IEEE VRW}, 
  title={Generative RGB-D Face Completion for Head-Mounted Display Removal}, 
  year={2021},
  volume={},
  number={},
  pages={109-116},
  doi={10.1109/VRW52623.2021.00028}}

@INPROCEEDINGS{Zhang2024REFA,
  author={Zhang, Qiang and Xiao, Tong and Habeeb, Haroun and Laich, Larissa and Bouaziz, Sofien and Snape, Patrick and Zhang, Wenjing and Cioffi, Matthew and Zhang, Peizhao and Pidlypenskyi, Pavel and Lin, Winnie and Ma, Luming and Wang, Mengjiao and Li, Kunpeng and Long, Chengjiang and Song, Steven and Prazak, Martin and Sjoholm, Alexander and Deogade, Ajinkya and Lee, Jaebong and Mangas, Julio Delgado and Aubel, Amaury},
  booktitle={Proc. IEEE CVPRW}, 
  title={REFA: Real-time Egocentric Facial Animations for Virtual Reality}, 
  year={2024},
  volume={},
  number={},
  pages={4793-4802},
  doi={10.1109/CVPRW63382.2024.00482}}

@misc{nierula_2025,
      title={Differential Physiological Responses to Proxemic and Facial Threats in Virtual Avatar Interactions}, 
      author={Birgit Nierula and Mustafa Tevfik Lafci and Anna Melnik and Mert Akgül and Farelle Toumaleu Siewe and Sebastian Bosse},
      year={2025},
      eprint={2508.10586},
      archivePrefix={arXiv},
      primaryClass={cs.HC},
      url={https://arxiv.org/abs/2508.10586}, 
}

@INPROCEEDINGS{llobera_2010,
  title = {Proxemics with Multiple Dynamic Characters in an Immersive Virtual Environment},
  author = {Llobera, Joan and Spanlang, Bernhard and Ruffini, Giulio and Slater, Mel},
  year = {2010},
  journal = {ACM Trans. Appl. Percept.},
  volume = {8},
  number = {1},
  pages = {1--12},
  doi = {10.1145/1857893.1857896},
}

@article{hu2025openface,
  title={OpenFace 3.0: A Lightweight Multitask System for Comprehensive Facial Behavior Analysis},
  author={Hu, Jiewen and Mathur, Leena and Liang, Paul Pu and Morency, Louis-Philippe},
  journal={arXiv preprint arXiv:2506.02891},
  year={2025}
}

@INPROCEEDINGS{Cohn2010ck+,
  author={Lucey, Patrick and Cohn, Jeffrey F. and Kanade, Takeo and Saragih, Jason and Ambadar, Zara and Matthews, Iain},
  booktitle={Proc. IEEE CVPRW}, 
  title={The Extended Cohn-Kanade Dataset (CK+): A complete dataset for action unit and emotion-specified expression}, 
  year={2010},
  volume={},
  number={},
  pages={94-101},
  doi={10.1109/CVPRW.2010.5543262}}

@article{kothe2025lab,
  title={The Lab Streaming Layer for Synchronized Multimodal Recording},
  author={Kothe, Christian and Shirazi, Seyed Yahya and Stenner, Tristan and Medine, David
          and Boulay, Chadwick and Grivich, Matthew I. and Artoni, Fiorenzo and Mullen, Tim
          and Delorme, Arnaud and Makeig, Scott},
  journal={Imaging Neuroscience},
  volume={3},
  pages={IMAG.a.136},
  year={2025},
  publisher={MIT Press},
  doi={10.1162/IMAG.a.136},
  url={https://doi.org/10.1162/IMAG.a.136},
  note={Open Access}
}

@misc{YOLOPose,
      title={YOLO-Pose: Enhancing YOLO for Multi Person Pose Estimation Using Object Keypoint Similarity Loss}, 
      author={Debapriya Maji and Soyeb Nagori and Manu Mathew and Deepak Poddar},
      year={2022},
      eprint={2204.06806},
      archivePrefix={arXiv},
      primaryClass={cs.CV},
      url={https://arxiv.org/abs/2204.06806}, 
}

@software{yolo11_ultralytics,
  author = {Glenn Jocher and Jing Qiu},
  title = {Ultralytics YOLO11},
  version = {11.0.0},
  year = {2024},
  url = {https://github.com/ultralytics/ultralytics},
  orcid = {0000-0001-5950-6979, 0000-0002-7603-6750, 0000-0003-3783-7069},
  license = {AGPL-3.0}
}

@article{fernandez-alcantara_2025,
    author = {{Fern{\'a}ndez-Alc{\'a}ntara}, Manuel and Escribano, Silvia and {Juli{\'a}-Sanchis}, Roc{\'i}o and {Castillo-L{\'o}pez}, Ana and {P{\'e}rez-Manzano}, Antonio and Macur, M and {Kalender-Smajlovi{\'c}}, Sedina and {Garc{\'i}a-Sanju{\'a}n}, Sof{\'i}a and {Caba{\~n}ero-Mart{\'i}nez}, Mar{\'i}a Jos{\'e}},
    title = {Virtual {{Simulation Tools}} for {{Communication Skills Training}} in {{Health Care Professionals}}: {{Literature Review}}},
    journal = {JMIR Medical Education},
    year = {2025},
    volume = {11},
    pages = {e63082-e63082},
    issn = {2369-3762},
    doi = {10.2196/63082},
}

@article{murtinger_2024,
    author = {Murtinger, Markus and Uhl, Jakob Carl and Atzm{\"u}ller, Lisa Maria and Regal, Georg and Roither, Michael},
    title = {Sound of the {{Police}}---{{Virtual Reality Training}} for {{Police Communication}} for {{High-Stress Operations}}},
    journal = {Multimodal Technologies and Interaction},
    year = {2024},
    volume = {8},
    number = {6},
    pages = {46},
    doi = {10.3390/mti8060046}
}

@article{sagliano_2022,
    author = {Sagliano, Laura and Ponari, Marta and Conson, Massimiliano and Trojano, Luigi},
    title = {Editorial: {{The}} Interpersonal Effects of Emotions: {{The}} Influence of Facial Expressions on Social Interactions},
    journal = {Frontiers in Psychology},
    year = {2022},
    volume = {13},
    pages = {1074216},
    doi = {10.3389/fpsyg.2022.1074216},
}

@inbook{CLAHE,
author = {Zuiderveld, Karel},
title = {Contrast limited adaptive histogram equalization},
year = {1994},
isbn = {0123361559},
publisher = {Academic Press Professional, Inc.},
address = {USA},
booktitle = {Graphics Gems IV},
pages = {474–485},
numpages = {12}
}

@inproceedings{Lianghui24_ViM,
author = {Zhu, Lianghui and Liao, Bencheng and Zhang, Qian and Wang, Xinlong and Liu, Wenyu and Wang, Xinggang},
title = {Vision mamba: efficient visual representation learning with bidirectional state space model},
year = {2024},
publisher = {JMLR.org},
booktitle = {Proceedings of the 41st International Conference on Machine Learning},
articleno = {2584},
numpages = {14},
location = {Vienna, Austria},
series = {ICML'24}
}

@article{dosovitskiy2020image,
  title={An image is worth 16x16 words: Transformers for image recognition at scale},
  author={Dosovitskiy, Alexey and Beyer, Lucas and Kolesnikov, Alexander and Weissenborn, Dirk and Zhai, Xiaohua and Unterthiner, Thomas and Dehghani, Mostafa and Minderer, Matthias and Heigold, Georg and Gelly, Sylvain and others},
  journal={arXiv preprint arXiv:2010.11929},
  year={2020}
}

@misc{qin2024mobilenetv4,
      title={MobileNetV4 -- Universal Models for the Mobile Ecosystem}, 
      author={Danfeng Qin and Chas Leichner and Manolis Delakis and Marco Fornoni and Shixin Luo and Fan Yang and Weijun Wang and Colby Banbury and Chengxi Ye and Berkin Akin and Vaibhav Aggarwal and Tenghui Zhu and Daniele Moro and Andrew Howard},
      year={2024},
      eprint={2404.10518},
      archivePrefix={arXiv},
      primaryClass={cs.CV}
}

@book{ekman1978manual,
  title = {Manual for the Facial Action Coding System},
  author = {Ekman, Paul and Friesen, Wallace V},
  year = {1978},
  publisher = {Consulting Psychologists Press}
}

@inproceedings{ortmann2023facial,
  title={Facial emotion recognition in immersive virtual reality: A systematic literature review},
  author={Ortmann, Thorben and Wang, Qi and Putzar, Larissa},
  booktitle={Proceedings of the 16th International Conference on PErvasive Technologies Related to Assistive Environments},
  pages={77--82},
  year={2023}
}

@article{williams_2023,
  title = {Developing and {{Validating}} a {{Facial Emotion Recognition Task With Graded Intensity}}},
  author = {Williams, Trevor F. and Vehabovic, Niko and Simms, Leonard J.},
  year = 2023,
  journal = {Assessment},
  volume = {30},
  number = {3},
  pages = {761--781},
  doi = {10.1177/10731911211068084},
}

@article{weber2021_vrinference,
  title = {A {{Structured Approach}} to {{Test}} the {{Signal Quality}} of {{Electroencephalography Measurements During Use}} of {{Head-Mounted Displays}} for {{Virtual Reality Applications}}},
  author = {Weber, Desir{\'e}e and Hertweck, Stephan and Alwanni, Hisham and Fiederer, Lukas D. J. and Wang, Xi and Unruh, Fabian and Fischbach, Martin and Latoschik, Marc Erich and Ball, Tonio},
  year = 2021,
  journal = {Frontiers in Neuroscience},
  volume = {15},
  pages = {733673},
  doi = {10.3389/fnins.2021.733673},
}

@article{dfg_2025,
  title = {{Guidelines for Safeguarding Good Research Practice. Code of Conduct}},
  author = {Deutsche Forschungsgemeinschaft},
  year = 2025,
  publisher = {Deutsche Forschungsgemeinschaft},
  doi = {10.5281/ZENODO.3923601},
}

@article{barker_2025,
  title = {Ethical {{Considerations}} in {{Emotion Recognition Research}}},
  author = {Barker, Darlene and Tippireddy, Mukesh Kumar Reddy and Farhan, Ali and Ahmed, Bilal},
  year = 2025,
  journal = {Psychology International},
  volume = {7},
  number = {2},
  pages = {43},
  doi = {10.3390/psycholint7020043},
}

\end{document}